\documentclass{article} %
\usepackage{iclr2023_conference,times}

\usepackage[english]{babel}

\usepackage{amsmath,amsfonts,bm}

\def\eqref#1{equation~\ref{#1}}

\def\1{\bm{1}}

\def\vh{{\bm{h}}}

\def\vu{{\bm{u}}}
\def\vv{{\bm{v}}}
\def\vw{{\bm{w}}}
\def\vx{{\bm{x}}}

\def\vz{{\bm{z}}}

\def\mA{{\bm{A}}}

\def\mH{{\bm{H}}}

\def\mX{{\bm{X}}}
\def\mY{{\bm{Y}}}
\def\mZ{{\bm{Z}}}

\DeclareMathAlphabet{\mathsfit}{\encodingdefault}{\sfdefault}{m}{sl}
\SetMathAlphabet{\mathsfit}{bold}{\encodingdefault}{\sfdefault}{bx}{n}

\def\gA{{\mathcal{A}}}

\def\gE{{\mathcal{E}}}

\def\gL{{\mathcal{L}}}

\def\gV{{\mathcal{V}}}

\newcommand{\E}{\mathbb{E}}

\newcommand{\R}{\mathbb{R}}

\long\def\IfNoTokA #1\IfNoTokB \IfNoTokB {}
\long\def\IfNoTokB \IfNoTokC #1#2{#1}    
\long\def\IfNoTokC           #1#2{#2}%
\makeatletter
\newcommand*{\tp}{%
  {\mathpalette\@transpose{}}%
}
\newcommand*{\@transpose}[2]{%
  \raisebox{\depth}{$\m@th#1\intercal$}%
}
\makeatother

\long\def\IfNoTokens #1{\IfNoTokA\IfNoTokB #1\IfNoTokB\IfNoTokB\IfNoTokC }

\definecolor{aqua}{rgb}{0.0, 1.0, 1.0}

\newcommand{\revAdd}[1]{{#1}}

\newcommand{\methodname}{{\small\textsf{T-BGRL}}\xspace}

\newcommand{\wtodo}[1]{{\textcolor{blue}{\textbf{[\IfNoTokens{#1}{TODO}{TODO: #1}]}}}}

\newcommand{\ms}[2]{{#1\tiny{$\pm$#2}}}

\newcommand{\dset}[1]{\texttt{#1}}

\newcommand{\tDist}{\textsc{Dist}}
\newcommand{\encoder}{\textsc{Enc}}
\newcommand{\decoder}{\textsc{Dec}}
\newcommand{\Enc}{\encoder}
\newcommand{\Dec}{\decoder}
\newcommand{\oEnc}{\encoder_\theta}
\newcommand{\tEnc}{\encoder_\phi}
\newcommand{\paset}{\gA^{+}}

\newcommand{\posaug}{\textsc{Aug}^+}
\newcommand{\negaug}{\textsc{Aug}^-}
\newcommand{\oracle}{\textsc{Orc}}
\newcommand{\pred}{\textsc{Pred}}
\newcommand{\neigh}{\textsc{Neigh}}
\newcommand{\maxPair}{(\mA, \mX)}
\newcommand{\pMaxPair}[1]{(\tilde{\mA}^{(#1)}, \tilde{\mX}^{(#1)})}
\newcommand{\nMaxPair}{(\check{\mA}, \check{\mX})}

\usepackage{listings}
\usepackage{enumitem}
\usepackage[utf8]{inputenc} %
\usepackage[T1]{fontenc}    %
\usepackage{url}            %
\usepackage{booktabs}       %
\usepackage{amsfonts,amsmath,amssymb,amsthm}
\usepackage{nicefrac}       %
\usepackage{microtype}      %
\usepackage{xcolor}         %
\usepackage{footnote}
\usepackage{duckuments}     %

\usepackage{bm}
\usepackage[utf8]{inputenc} %
\usepackage[T1]{fontenc}    %
\usepackage{hyperref}       %
\usepackage{url,hyphenat}
\usepackage{caption,subfigure,graphicx,epstopdf,verbatim,wrapfig}
\usepackage{makecell}
\usepackage{float}
\usepackage{algorithmic}
\usepackage{pifont}

\usepackage{bbding} 
\usepackage{soul}
\usepackage{threeparttable}
\usepackage{multirow,multicol}
\usepackage{adjustbox}
\usepackage{tikzducks}
\usepackage{placeins}
\usepackage{totcount,etoolbox}
\usepackage[vlined,boxed,commentsnumbered,linesnumbered,ruled]{algorithm2e}

\usepackage[capitalize,noabbrev,nameinlink]{cleveref}
\usepackage{xspace}

\usepackage[textsize=tiny]{todonotes}
\usepackage{marginnote}

\definecolor{mydarkblue}{rgb}{0,0.08,0.45}
\definecolor{myblue}{HTML}{3b75c3}
\definecolor{myred}{HTML}{E33222}
\definecolor{mygreen}{HTML}{438773}
\definecolor{mymaroon}{RGB}{142,27,19}
\definecolor{maroon}{HTML}{800000}
\definecolor{mycite}{cmyk}{0.55,1,0,0.15}
\definecolor{codeblue}{rgb}{0.25,0.5,0.5}
\definecolor{codekw}{rgb}{0.85, 0.18, 0.50}
\definecolor{codegreen}{rgb}{0,0.6,0}
\definecolor{codegray}{rgb}{0.5,0.5,0.5}
\definecolor{codepurple}{rgb}{0.58,0,0.82}
\definecolor{backcolour}{rgb}{0.95,0.95,0.92}
\hypersetup{
    colorlinks=true,
    citecolor=mydarkblue,
    linkcolor=codekw,
    urlcolor=maroon
          }

\title{Link Prediction with Non-Contrastive \\Learning}

\author{
William Shiao$^{1}$\thanks{Work done while interning at Snap Inc.}, Zhichun Guo$^{2}\footnotemark[1]$, Tong Zhao$^{3}$, \\ 
\ \textbf{Evangelos E. Papalexakis$^{1}$, Yozen Liu$^{3}$, Neil Shah$^{3}$}\\
$^{1}$University of California, Riverside\ \
$^{2}$University of Notre Dame \ \
$^{3}$Snap Inc.\\
$^{1}$\texttt{\{wshia002,epapalex\}@ucr.edu},\ $^{2}$\texttt{zguo5@nd.edu}, \ $^{3}$\texttt{\{tzhao,yliu2,nshah\}@snap.com}\\
}

\iclrfinalcopy %

\begin{document}
\theoremstyle{definition}
\newtheorem{definition}{Definition}[section]

\maketitle

\begin{abstract}
Graph neural networks (GNNs) are prominent in the graph machine learning domain, owing to their strong performance across various tasks.  A recent focal area is the space of graph self-supervised learning (SSL), which aims to derive useful node representations without labeled data. 
Notably, many state-of-the-art graph SSL approaches are \textit{contrastive} methods, which use a combination of positive and negative samples to learn node representations.  Owing to challenges in negative sampling (slowness and model sensitivity), recent literature introduced \textit{non-contrastive} methods, which instead only use positive samples.  Though such methods have shown promising performance in node-level tasks, their suitability for link prediction tasks, which are concerned with predicting link existence between pairs of nodes, and have broad applicability to recommendation systems contexts, is yet unexplored.
In this work, we extensively evaluate the performance of existing non-contrastive methods for link prediction in both transductive and inductive settings. While most existing non-contrastive methods perform poorly overall, we find that, surprisingly, BGRL generally performs well in transductive settings. However, it performs poorly in the more realistic inductive settings where the model has to generalize to links to/from unseen nodes. We find that non-contrastive models tend to overfit to the training graph and use this analysis to propose \methodname, a novel non-contrastive framework that incorporates cheap corruptions to improve the generalization ability of the model. This simple modification strongly improves inductive performance in \textbf{5/6} of our datasets, with up to a \textbf{120\%} improvement in Hits@50---all with comparable speed to other non-contrastive baselines, and up to \textbf{14\texttimes{}} faster than the best-performing contrastive baseline. Our work imparts interesting findings about non-contrastive learning for link prediction and paves the way for future researchers to further expand upon this area.

\end{abstract}

\section{Introduction}

Graph neural networks (GNNs) are ubiquitously used modeling tools for relational graph data, with widespread applications in chemistry \citep{chen2019graph,guo2021few,guo2022graph,liu2022graph}, forecasting and traffic prediction \citep{derrow2021eta, tang2020knowing}, recommendation systems~\citep{ying2018graph,he2020lightgcn,sankar2021graph, tang2022friend,fan2022graph}, graph generation~\citep{graphrnn,fan2019labeled,shiao2021adversarially}, and more.  Given significant challenges in obtaining labeled data, one particularly exciting recent direction is the advent of graph self-supervised learning (SSL), which aims to learn representations useful for various downstream tasks without using explicit supervision besides available graph structure and node features \citep{grace, jin2021automated,bgrl, gbt}.

One prominent class of graph SSL approaches are contrastive methods~\citep{jin2020self}. 
These methods typically utilize contrastive losses such as InfoNCE~\citep{oord2018representation} or margin-based losses~\citep{ying2018graph} between node and negative sample representations. %
However, such methods usually require either many negative samples~\citep{mvgrl} or carefully chosen ones~\citep{ying2018graph,understanding_neg_sampling}, where the first one results with quadratic number of in-batch comparisons, and the latter is especially expensive on graphs since we often store the sparse adjacency matrix instead of its dense complement~\citep{bgrl, gbt}. 
These drawbacks motivated the development of non-contrastive methods \citep{bgrl,gbt,ccaSSG,selfgnn}, based on advances in the image domain \citep{byol,simsiam,simclr}, which do not require negative samples and solely rely on augmentations. This allows for a large speedup compared to their contrastive counterparts with strong performance \citep{gbt,ccaSSG}.

However, non-contrastive SSL methods are typically evaluated on node-level tasks, %
which is a more direct analog of image classification in the graph domain. In comparison, the link-level task (link prediction), which focuses on predicting link existence between pairs of nodes, is largely overlooked. This presents a critical gap in understanding: \emph{Are non-contrastive methods suitable for link prediction tasks? When do they (not) work, and why?} This gap presents a huge opportunity, since link prediction is a cornerstone in the recommendation systems community \citep{he2020lightgcn, zhang2019inductive, berg2017graph}.%

\textbf{Present Work.} To this end, our work first performs an extensive evaluation of non-contrastive SSL methods in link prediction contexts to discover the impact of different augmentations, architectures, and non-contrastive losses. We evaluate all of the (to the best of our knowledge) currently existing non-contrastive methods: CCA-SSG~\citep{ccaSSG}, \revAdd{Graph Barlow Twins} (GBT)~\citep{gbt}, and \revAdd{Bootstrapped Graph Latents} (BGRL)~\citep{bgrl} (which has the same design as the independently proposed SelfGNN~\citep{selfgnn}). We also compare these methods against a baseline end-to-end GCN~\citep{gcn} with cross-entropy loss, and two contrastive baselines: GRACE~\citep{grace}, and a GCN trained with max-margin loss~\citep{pinSAGE}. We evaluate the methods in the transductive setting and find that BGRL~\citep{bgrl} greatly outperforms not only the other non-contrastive methods, but also GRACE---a strong augmentation-based contrastive model for node classification. Surprisingly, BGRL even performs on-par with a margin-loss GCN (with the exception of 2/6 datasets). %
However, in the more realistic inductive setting, which considers prediction between new edges and nodes at inference time, we observe a huge gap in performance between BGRL and a margin-loss GCN (ML-GCN). %
Upon investigation, we find that BGRL is unable to sufficiently push apart the representations of negative links from positive links when new nodes are introduced, owing to a form of overfitting. %
To address this, we propose \methodname, %
 a novel non-contrastive method which uses a corruption function to generate cheap ``negative''  samples---without performing the expensive negative sampling step of contrastive methods. We show that it greatly reduces overfitting tendencies, and outperforms existing non-contrastive methods across 5/6 datasets on the inductive setting. We also show that it maintains comparable speed with BGRL, and is 14\texttimes{} faster than the margin-loss GCN on the \dset{Coauthor-Physics} dataset.

\textbf{Main Contributions}. In short, our main contributions are as follows:
\begin{itemize}[leftmargin=6mm,itemsep=2pt,parsep=4pt,topsep=0pt,partopsep=2pt]
    \item To the best of our knowledge, this is the first work to explore link prediction with non-contrastive SSL methods.
    \item We show that, perhaps surprisingly, BGRL (an existing non-contrastive model) works well in the transductive link prediction, with performance at par with contrastive baselines, implicitly behaving similarly to other contrastive models in pushing apart positive and negative node pairs. %
    \item We show that non-contrastive SSL models underperform their contrastive counterparts in the inductive setting, and notice that they generalize poorly due to a lack of negative examples.
    \item Equipped with this understanding, we propose \methodname{}, a novel non-contrastive method that uses cheap ``negative'' samples to improve generalization. \methodname is simple to implement, very efficient when compared to contrastive methods, and improves on BGRL's inductive performance in 5/6 datasets, making it at or above par with the best contrastive baselines.
\end{itemize}

\section{Preliminaries}
\label{sec:prelims}
\textbf{Notation.} We denote a graph as $G = (\gV, \gE)$, where $\gV$ is the set of $n$ nodes (i.e., $n = |\gV|$) and $\gE \subseteq \gV \times \gV$ be the set of edges. Let the node-wise feature matrix be denoted by $\mX \in \R^{n \times f}$, where $f$ is the number of raw features, and its $i$-th row $\vx_i$ is the feature vector for the $i$-th node. Let $\mA \in \{0, 1\}^{n \times n}$ denote the binary adjacency matrix. We denote the graph's learned node representations as $\mH \in \R^{n \times d}$, where $d$ is the size of latent dimension, and $\vh_i$ is the representation for the $i$-th node. Let $\mY \in \{0, 1\}^{n \times n}$ be the desired output for link prediction,
as $\gE$ and $\mA$ may have validation and test edges masked off. Similarly, let $\hat{\mY} \in \{0, 1\}^{n \times n}$ be the output predicted by the decoder for link prediction. Let $\oracle$ be a perfect oracle function for our link prediction task, i.e., $\oracle{}(\mA, \mX) = \mY$. Let $\neigh (u) = \{ v\ |\ (u, v) \in\gE \lor (v,u) \in \gE \}$. Note that we use the terms ``embedding'' and ``representation'' interchangeably in this work.

\textbf{GNNs for Link Prediction.} Many new approaches have also been developed with the recent advent of graph neural networks (GNNs). A predominant paradigm is the use of node-embedding-based methods \citep{graphsage, berg2017graph, ying2018graph, zhao2022learning}. %
Node-embedding-based methods typically consist of an encoder $\mH = \Enc(\mA, \mX)$ and a decoder $\Dec(\mH)$. The encoder model is typically a message-passing based Graph Neural Network (GNN)~\citep{gcn,graphsage,zhang2020deep}. The message-passing iterations of a GNN for a node $u$ can be described as follows:
\begin{equation}
    \vh_u^{(k+1)} = \textsc{Update}^{(k)} \left(
        \vh_u^{(k)}, \textsc{Aggregate}^{(k)} (
            \{  \vh_v^{(k)}, \forall v \in \neigh (u) \}
        )
    \right)
\end{equation}%
where \textsc{Update} and \textsc{Aggregate} are differentiable functions, and $\vh^{(0)}_u = \vx_u$. The decoder model is usually an inner product or MLP applied on a concatenation of Hadamard product of the source and target learned node representations \citep{rendle2020neural, wang2021pairwise}. 

\textbf{Graph SSL.} %
Below, we define a few terms used throughout our work which helps set the context for our discussion.%

\begin{definition}[Augmentation]%
An augmentation $\posaug{}$ is a label-preserving random transformation function $\posaug{}:(\mA, \mX){\rightarrow}(\tilde{\mA}, \tilde{\mX})$ that does not change the oracle's expected value: $\E{[ \oracle{}(\posaug(\mA, \mX)) ]} = \mY$.
\end{definition}

\begin{definition}[Corruption]%
A corruption $\negaug{}$ is a label-altering random transformation $\negaug{}:(\mA, \mX) \rightarrow (\check{\mA}, \check{\mX})$ that changes the oracle's expected value: $\E{[ \oracle{}(\negaug(\mA, \mX)) ]} \neq \mY$.\footnote{Note that the definition of these functions are different from the corruption functions in \citet{grace} (which we define as \textit{augmentations}) and are instead similar to the corruption functions in \citet{dgi}.}
\end{definition}

\begin{definition}[Contrastive Learning]
Contrastive methods select anchor samples (e.g. nodes) and then compare those samples to both \textit{positive} samples (e.g. neighbors) and \textit{negative} samples (e.g. non-neighbors) relative to those anchor samples.
\end{definition}%

\begin{definition}[Non-Contrastive Learning]
Non-contrastive methods select anchor samples, but only compare those samples to variants of themselves, without leveraging other samples in the dataset.%
\end{definition}

\textbf{BGRL.} %
While we examine the performance of all of the non-contrastive graph models, we focus our detailed analysis exclusively on BGRL\footnote{Self-GNN \citep{selfgnn}, which was published independently, also shares the same architecture. As such, we refer to these two methods as BGRL.} \citep{bgrl} due to its superior performance in link prediction when compared to GBT \citep{gbt} and CCA-SSG \citep{ccaSSG}. BGRL consists of two encoders, one of which is referred to as the \textit{online} encoder $\oEnc$; the other is referred to as the \textit{target} encoder $\tEnc$. BGRL also incorporates a predictor $\pred$ (typically a MLP) and two sets of augmentations: $\paset_1, \paset_2$. A single training step for BGRL is as follows: (a) we apply these augmentations: $\pMaxPair{1} = \posaug_1 \maxPair;\ \pMaxPair{2} = \posaug_2 \maxPair$. (b) we perform forward propagation $\mH = \Enc \pMaxPair{1}; \mH_2 = \Enc \pMaxPair{2}$. %
(c) we pass the output through the predictor $\mZ = \pred(\mH_1)$. (d) we use the mean pairwise cosine distance of $\mZ$ and $\mH_2$ as the loss (see Eqn.~\ref{eqn:bgrl_loss}). (e) $\oEnc$ is updated via backpropagation and $\tEnc$ is updated via exponential moving average (EMA) from $\oEnc$. %
The BGRL loss is as follows:%
\begin{equation}
\label{eqn:bgrl_loss}
    \gL_{\mathrm{BGRL}} = - \frac{2}{n} \sum^{n-1}_{i=0} \frac{\tilde{\vz}_i \cdot \vh_i^{(2)}}{||\tilde{\vz}_i||\ ||\vh_i^{(2)}||}
\end{equation}

\revAdd{In the next section, we evaluate BGRL and other non-contrastive link prediction methods against contrastive baselines.}

\section{Do Non-Contrastive Learning Methods Perform Well on Link Prediction Tasks?}
\vspace{-0.1in}
Several non-contrastive methods have been proposed and have shown effectiveness in node classification~\citep{selfgnn,bgrl,ccaSSG,gbt}. \revAdd{However,} none \revAdd{of these methods} evaluate or target link prediction tasks. We thus aim to answer the following questions: First, how well do these methods work for link prediction compared to existing contrastive/end-to-end baselines? Second, do they work equally well in  both transductive and inductive settings? Finally, if they do work, why; if not, why not? %

\textbf{Differences from Node Classification.} Link prediction differs from node classification in several key aspects. First, we must consider the embedding of both the source and destination nodes. Second, we have a much larger set of candidates for the same graph---$O(n^2)$ instead of $O(n)$. Finally, in real applications, link prediction is usually treated as a ranking problem, where we want positive links to be ranked higher than negative links, rather than as a classification problem, e.g. in recommendation systems, where we want to retrieve the top-$k$ most likely links \citep{recsysPerf,newStratsRecommendation}. We discuss this in more detail in \cref{subsec:evaluation} below. 
Given these differences, it is unclear if methods performing well on node classification naturally perform well on link prediction tasks.

\textbf{Ideal Link Prediction.} %
What does it mean to perform well on link prediction? We clarify this point here. For some nodes $u,v,w \in \gV$, let  $(u,v) \in \gE$ and $(u, w) \not \in \gE$. Then, an ideal encoder for link prediction would have $\tDist(\vh_u, \vh_v) < \tDist(\vh_u,\vh_w)$ for some distance function $\tDist$. This idea is the core motivation behind margin-loss-based models~\citep{pinSAGE,graphsage}.

\subsection{Evaluation}
\label{subsec:evaluation}

\textbf{Datasets.} We use datasets from three different domains: citation networks, co-authorship networks, and co-purchase networks. We use the \dset{Cora} and \dset{Citeseer} citation networks~\citep{coraCiteseer}, the \dset{Coauthor-CS} and \dset{Coauthor-Physics} co-authorship networks, and the \dset{Amazon-Computers} and \dset{Amazon-Photos} co-purchase networks~\citep{amazonComputersPhotos}. We include dataset statistics in \cref{subsec:dset_stats}.

\textbf{Metric.} Following work in the heterogeneous information network~\citep{pme}, knowledge-graph~\citep{learningEntityKGC}, and recommendation systems~\citep{recsysPerf,newStratsRecommendation} communities, we choose to use Hits@$k$ over AUC-ROC metrics, since we often empirically prioritize ranking candidate links from a selected node context (e.g. ranking the probability that user $A$ will buy item $B$, $C$, or $D$), as opposed to arbitrarily ranking a randomly chosen positive over negative link (e.g. ranking whether the probability that user $A$ buys item $B$ is more likely than user $C$ does not buy item $D$). 
 We report Hits@50 ($k = 50$) to strike a balance between the smaller datasets like \dset{Cora} and the larger datasets like $\dset{Coauthor-Physics}$. However, for completeness of the evaluation, %
 we also include AUC-ROC results in \cref{subsec:auc_res}.

\textbf{Decoder.} Since our goal is to evaluate the performance of the encoder, we use the same decoder for all of our experiments across all of the methods. \revAdd{The choice of decoder has also been previously studied~\citep{wang2021pairwise,wang2022flashlight}, so we use the best-performing decoder - a Hadamard product MLP.} For a candidate link $(u, v)$, we have $\hat{\mY} = \Dec( \vh_u * \vh_v )$ where $*$ represents the Hadamard product, and \Dec{} is a two-layer MLP (with 256 hidden units) followed by a sigmoid.
For the self-supervised methods, we first train the encoder and freeze its weights before training the decoder. As a contextual baseline, we also report results on an end-to-end GCN (E2E-GCN), for which we train the encoder and decoder jointly, backpropagating a binary cross-entropy loss on link existence.

\begin{table}
\vspace{-0.2in}
\centering
\caption{Transductive performance of different link prediction methods. We \textbf{bold} the best-performing method and \underline{underline} the second-best method for each dataset. BGRL consistently outperforms other non-contrastive methods and GRACE, and also outperforms ML-GCN, on 3/6 datasets.%
}
\label{tab:transductive}
\resizebox{\linewidth}{!}{%
\begin{tabular}{l||c||cc|ccc} 
\toprule
&\multicolumn{1}{c||}{\textbf{End-To-End}} & \multicolumn{2}{c|}{\textbf{Contrastive}} & \multicolumn{3}{c}{\textbf{Non-Contrastive}} \\
\midrule
\textbf{Dataset} & \textbf{E2E-GCN%
} & \textbf{ML-GCN} & \textbf{GRACE} & \textbf{CCA-SSG} & \textbf{GBT} & \textbf{BGRL} \\ 
\midrule
\dset{Cora} &\ms{\textbf{0.816}}{0.013} &\ms{\underline{0.815}}{0.002} &\ms{0.686}{0.056} &\ms{0.348}{0.091} & \ms{0.460}{0.149} &\ms{0.792}{0.015} \\
\dset{Citeseer} &\ms{\underline{0.822}}{0.017} &\ms{0.771}{0.020} &\ms{0.707}{0.068} &\ms{0.249}{0.168} &\ms{0.472}{0.196} &\ms{\textbf{0.858}}{0.020} \\
\dset{Amazon-Photos} &\ms{\textbf{0.642}}{0.029} & \ms{0.430}{0.032} &\ms{0.486}{0.025} &\ms{0.369}{0.013} &\ms{0.434}{0.038} &\ms{\underline{0.562}}{0.013} \\
\dset{Amazon-Computers} &\ms{\textbf{0.426}}{0.036} & \ms{0.320}{0.060} & \ \ \ms{0.240}{0.027} &\ms{0.201}{0.032} &\ms{0.258}{0.008} &\ms{\underline{0.346}}{0.018} \\
\dset{Coauthor-CS} &\ms{\underline{0.762}}{0.010} &\ms{\textbf{0.787}}{0.011} &\ms{0.456}{0.066} &\ms{0.229}{0.018} &\ms{0.298}{0.033} &\ms{0.515}{0.016} \\
\dset{Coauthor-Physics} &\ms{\underline{0.798}}{0.018} & \ms{\textbf{0.810}}{0.003} &OOM &\ms{0.157}{0.009} &\ms{0.187}{0.011} &\ms{0.476}{0.015} \\
\bottomrule
\end{tabular}
}
\end{table}

\begin{figure}[t!]
    \centering
    {\includegraphics[width=0.48\textwidth]{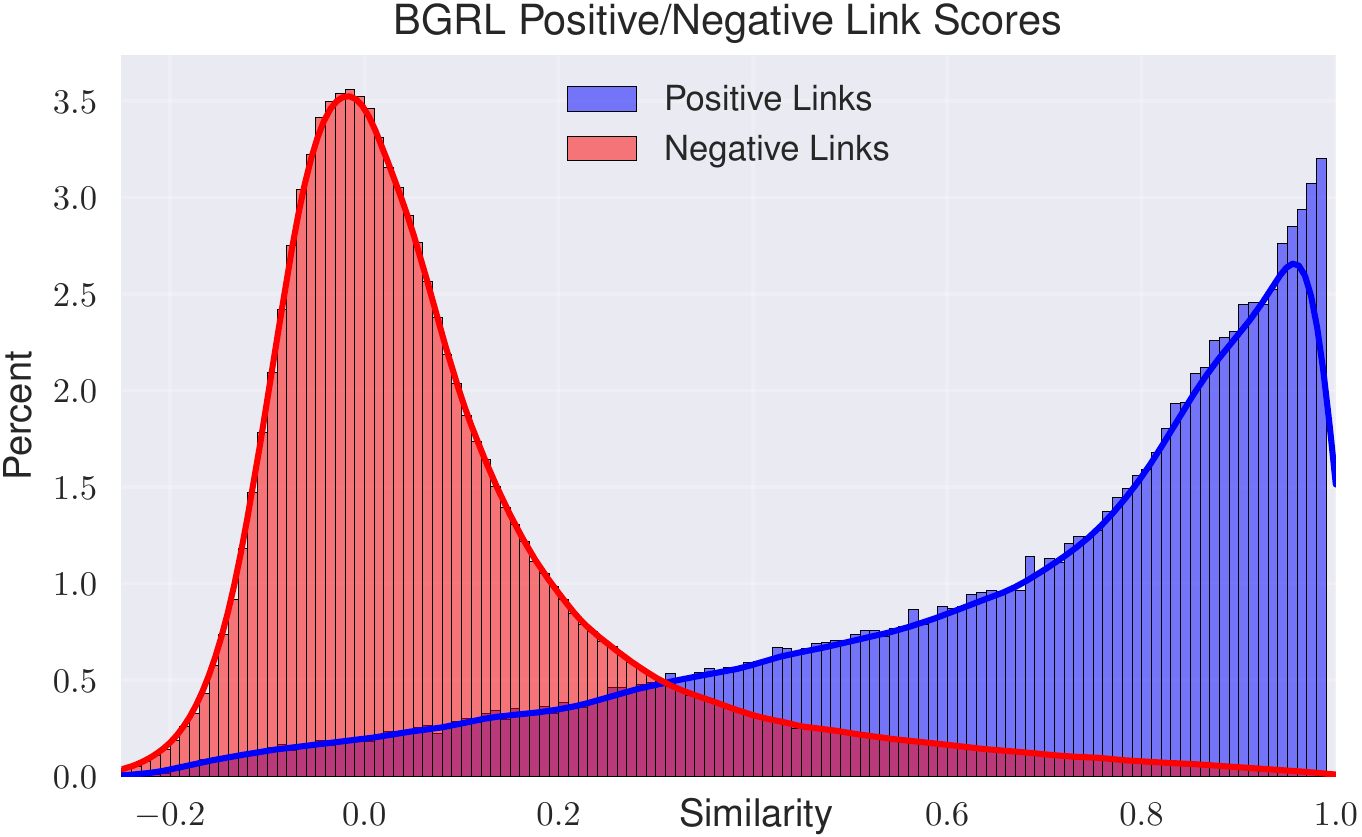}}
    \hfill
    {\includegraphics[width=0.48\textwidth]{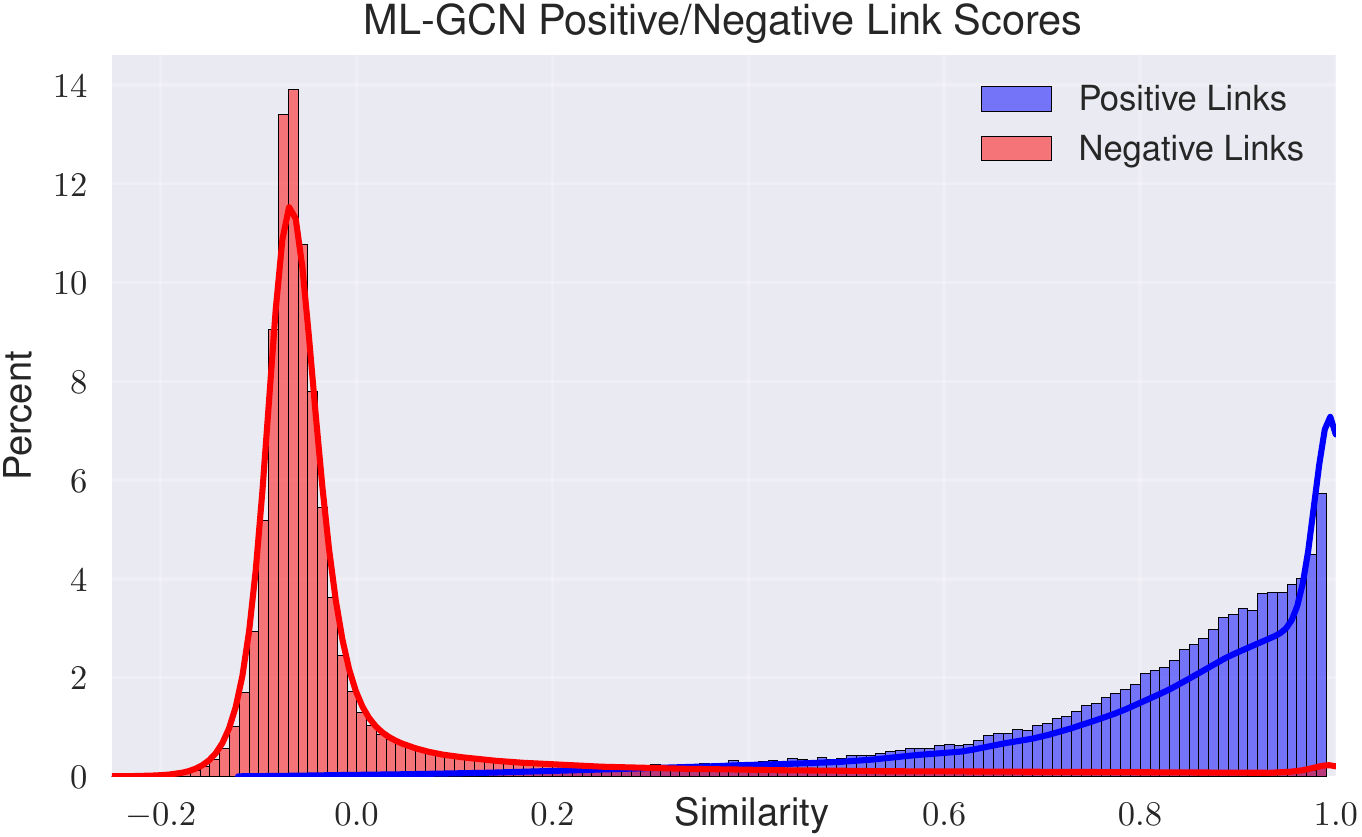}}
    \caption{\revAdd{These plots show similarities between node embeddings.} \textbf{Left:} distribution of positive/negative \revAdd{link similarities} for BGRL. \textbf{Right:} distribution of positive/negative \revAdd{link similarities} for ML-GCN. We can see that while they behave similarly, the ML-GCN does a better job of ensuring that positive/negative links are well separated. These scores are computed on \dset{Amazon-Photos}.}
    \label{fig:trans_dists}
\end{figure}

\subsubsection{Transductive Evaluation}
\label{subsubsec:transductive_eval}
\textbf{Transductive Setting.} We first evaluate the performance of the methods in the transductive setting, where we train on $G_{train} = (\gV, \gE_{train})$ for $\gE_{train} \subset \gE$, validate our method on $G_{val} = (\gV, \gE_{val})$ for $\gE_{val} \subset (\gE - \gE_{train})$, and test on $G_{test} = (\gV, \gE_{test})$ for $\gE_{test} = \gE - \gE_{train} - \gE_{val}$. Note that the same nodes are present in training, validation, and testing. We also do not introduce any new edges during inference time---inference is performed on $\gE_{train}$.

\textbf{Results.} The results of our evaluation are shown in \cref{tab:transductive}. As expected, the \revAdd{end-to-end} GCN generally performs the best across all of the datasets. We also find that CCA-SSG and GBT similarly perform poorly relative to the other methods. This is intuitive, as neither method was designed for link prediction and were only evaluated for node classification in their respective papers. Surprisingly, however, BGRL outperforms the ML-GCN (the strongest contrastive baseline) on 3/6 of the datasets and performs similarly on 1 other (\dset{Cora}). It also outperforms GRACE across all of the datasets. 

\textbf{Understanding BGRL Performance.} Interestingly, we find that BGRL exhibits similar behavior to the ML-GCN on many datasets, despite the BGRL loss function (see \cref{eqn:bgrl_loss}) not explicitly optimizing for this. \revAdd{Relative to an anchor node $u$, we can express the max-margin loss of the ML-GCN as follows:}
\begin{equation}
    L(u) = \E_{v \sim \neigh(u)} \left[
        \E_{w \sim \gE - \neigh(u)}
            J(u, v, w)
    \right]
\end{equation}%
where $J(u, v, w)$ is the margin ranking loss for an anchor $u$, positive sample $v$, and negative $w$:%
\begin{equation}
    J(u, v, w) = \max \{ 0, \vh_u \cdot \vh_v - \vh_u \cdot \vh_w + \Delta \}
\end{equation} %
and $\Delta$ is a hyperparameter for the size of the margin. This seemingly explicitly optimizes for the aforementioned ideal link prediction behavior (anchor-aware ranking of positive over negative links). Despite these circumstances, \cref{fig:trans_dists} shows that both BGRL and ML-GCN both clearly separate positive and negative samples, although ML-GCN pushes them further apart. \revAdd{We provide some intuition on why this may occur in \cref{subsec:pulling_repr_together} below.}%

\textbf{Why Does BGRL Not Collapse?} \revAdd{The loss function for BGRL (see \cref{eqn:bgrl_loss}) is 0 when $h_i^{(2)}$ = 0 or $\tilde{\vz}_i$ = 0, i.e., the loss is minimized when the model produces all-zero outputs. While theoretically possible, this is clearly undesirable behavior since this does not result in useful embeddings. We refer to this case as model collapse.} It is not fully understood why non-contrastive models do not collapse, but there have been several reasons proposed in the image domain with both theoretical and empirical grounding. We discuss this more in \cref{subsec:why_no_collapse}. \revAdd{Consistent with the findings from \citet{bgrl}, we find that collapse does not occur in practice (with reasonable hyperparameter selection).}

\revAdd{\textbf{Conclusion.} We find that CCA-SSG and GBT generally perform poorly compared to contrastive baselines. Surprisingly, we find that BGRL generally performs well in the transductive setting by successfully separating positive and negative link distance distributions. However, this setting may not be representative of real-world problems. In the next section, we evaluate the methods in the more realistic inductive setting to see if this performance holds.}

\begin{savenotes}
{
\begin{table}[t!]
\centering
\small
\vspace{-0.25in}
\caption{Performance of various methods in the inductive setting. See \cref{subsubsec:inductive_eval} for an explanation of our inductive setting.  Although we do not introduce \methodname{} until \cref{sec:method}, we include the results here to save space.}
\label{tab:inductive}
\resizebox{\columnwidth}{!}{
\begin{tabular}{l||c||cc|cccc}
\toprule
 &\multicolumn{1}{c||}{\textbf{End-To-End}} & \multicolumn{2}{c|}{\textbf{Contrastive}} & \multicolumn{4}{c}{\textbf{Non-Contrastive}} \\
 \midrule
 Dataset                  & \textbf{E2E-GCN}   & \textbf{ML-GCN}     & \textbf{GRACE}      & \textbf{GBT}        & \textbf{CCA-SSG}    & \textbf{BGRL}       &  \textbf{T-BGRL} \\
\midrule
\midrule
\multicolumn{8}{c}{Overall}\\ \midrule
\dset{Cora}               & \ms{\underline{0.523}}{0.019}  &  \ms{0.490}{0.028}  &  \ms{0.448}{0.043}  &  \ms{0.135}{0.077}  &  \ms{0.120}{0.018}  &  \ms{0.324}{0.184}  &  \ms{\textbf{0.568}}{0.033} \\
\dset{Citeseer}           & \ms{\underline{0.621}}{0.034}  &  \ms{0.661}{0.036}  &  \ms{0.514}{0.053}  &  \ms{0.305}{0.026}  &  \ms{0.170}{0.071}  &  \ms{0.526}{0.055}  &  \ms{\textbf{0.727}}{0.027} \\
\dset{Coauthor-Cs}        & \ms{0.484}{0.048}  &  \ms{\textbf{0.572}}{0.037}  &  \ms{0.313}{0.017}  &  \ms{0.182}{0.025}  &  \ms{0.176}{0.013}  &  \ms{0.438}{0.025}  &  \ms{\underline{0.534}}{0.026} \\
\dset{Coauthor-Physics}   & \ms{0.386}{0.016}  &  \ms{\textbf{0.550}}{0.059}  &  OOM                &  \ms{0.112}{0.014}  &  \ms{0.037}{0.051}  &  \ms{0.439}{0.013}  &  \ms{\underline{0.463}}{0.023} \\
\dset{Amazon-Computers}   & \ms{0.179}{0.010}  &  \ms{\underline{0.279}}{0.044}  &  \ms{0.212}{0.057}  &  \ms{0.172}{0.015}  &  \ms{0.155}{0.013}  &  \ms{0.270}{0.034}  &  \ms{\textbf{0.312}}{0.027} \\
\dset{Amazon-Photos}      & \ms{0.420}{0.123}  &  \ms{\textbf{0.478}}{0.008}  &  \ms{0.262}{0.010}  &  \ms{0.289}{0.032}  &  \ms{0.182}{0.072}  &  \ms{\underline{0.460}}{0.023}  &  \ms{0.450}{0.017} \\
\midrule \midrule

\multicolumn{8}{c}{Performance on Observed-Observed Node Edges}\\ \midrule                                    
\dset{Cora}               & \ms{\underline{0.574}}{0.020}   &  \ms{0.490}{0.029}  &  \ms{0.557}{0.038}  &  \ms{0.149}{0.084}  &  \ms{0.124}{0.026}  &  \ms{0.345}{0.196}  &  \ms{\textbf{0.624}}{0.027} \\
\dset{Citeseer}           & \ms{0.610}{0.023}  &  \ms{\underline{0.621}}{0.021}  &  \ms{0.602}{0.050}  &  \ms{0.358}{0.031}  &  \ms{0.197}{0.082}  &  \ms{0.605}{0.045}  &  \ms{\textbf{0.768}}{0.021} \\
\dset{Coauthor-Cs}        & \ms{0.504}{0.047}  &  \ms{\textbf{0.591}}{0.034}  &  \ms{0.332}{0.018}  &  \ms{0.187}{0.023}  &  \ms{0.177}{0.013}  &  \ms{0.462}{0.025}  &  \ms{\underline{0.535}}{0.026} \\
\dset{Coauthor-Physics}   & \ms{0.390}{0.015}  &  \ms{\textbf{0.566}}{0.058}  &  OOM                &  \ms{0.117}{0.014}  &  \ms{0.039}{0.054}  &  \ms{0.445}{0.012}  &  \ms{\underline{0.469}}{0.023} \\
\dset{Amazon-Computers}   & \ms{0.177}{0.009}  &  \ms{\underline{0.278}}{0.044}  &  \ms{0.212}{0.059}  &  \ms{0.169}{0.016}  &  \ms{0.155}{0.014}  &  \ms{0.270}{0.034}  &  \ms{\textbf{0.313}}{0.027} \\
\dset{Amazon-Photos}      & \ms{0.418}{0.123}  &  \ms{\textbf{0.483}}{0.009}  &  \ms{0.265}{0.011}  &  \ms{0.295}{0.031}  &  \ms{0.185}{0.070}  &  \ms{\underline{0.467}}{0.023}  &  \ms{0.457}{0.015} \\
\midrule \midrule

\multicolumn{8}{c}{Performance on Observed-Unobserved Node Edges}\\ \midrule                                    
\dset{Cora}             & \ms{0.462}{0.023}  &  \ms{\underline{0.487}}{0.021}  &  \ms{0.367}{0.045}  &  \ms{0.128}{0.075}  &  \ms{0.115}{0.014}  &  \ms{0.309}{0.175}  &  \ms{\textbf{0.528}}{0.037} \\
\dset{Citeseer}         & \ms{0.645}{0.055}  &  \ms{\underline{0.705}}{0.039}  &  \ms{0.458}{0.063}  &  \ms{0.280}{0.024}  &  \ms{0.148}{0.067}  &  \ms{0.487}{0.064}  &  \ms{\textbf{0.708}}{0.034} \\
\dset{Coauthor-Cs}      & \ms{0.459}{0.049}  &  \ms{\textbf{0.545}}{0.042}  &  \ms{0.284}{0.017}  &  \ms{0.175}{0.026}  &  \ms{0.177}{0.013}  &  \ms{0.402}{0.025}  &  \ms{\underline{0.536}}{0.027} \\
\dset{Coauthor-Physics} & \ms{0.379}{0.019}  &  \ms{\textbf{0.525}}{0.058}  &  OOM                &  \ms{0.106}{0.013}  &  \ms{0.035}{0.048}  &  \ms{0.429}{0.013}  &  \ms{\underline{0.455}}{0.022} \\
\dset{Amazon-Computers} & \ms{0.183}{0.010}  &  \ms{\underline{0.281}}{0.045}  &  \ms{0.213}{0.056}  &  \ms{0.177}{0.014}  &  \ms{0.155}{0.011}  &  \ms{0.270}{0.034}  &  \ms{\textbf{0.312}}{0.027} \\
\dset{Amazon-Photos}    & \ms{0.424}{0.123}  &  \ms{\textbf{0.470}}{0.007}  &  \ms{0.258}{0.011}  &  \ms{0.279}{0.032}  &  \ms{0.178}{0.076}  &  \ms{\underline{0.449}}{0.022}  &  \ms{0.439}{0.021} \\
\midrule \midrule

\multicolumn{8}{c}{Performance on Unobserved-Unobserved Node Edges}\\ \midrule                                    
\dset{Cora}             & \ms{0.239}{0.027}  &  \ms{\textbf{0.507}}{0.063}  &  \ms{0.252}{0.066}  &  \ms{0.100}{0.076}  &  \ms{0.125}{0.020}  &  \ms{0.287}{0.164}  &  \ms{\underline{0.463}}{0.065} \\
\dset{Citeseer}         & \ms{\underline{0.595}}{0.073}  &  \ms{\textbf{0.681}}{0.101}  &  \ms{0.287}{0.039}  &  \ms{0.137}{0.019}  &  \ms{0.126}{0.043}  &  \ms{0.271}{0.078}  &  \ms{\underline{0.595}}{0.045} \\
\dset{Coauthor-Cs}      & \ms{0.372}{0.043}  &  \ms{\underline{0.483}}{0.046}  &  \ms{0.230}{0.019}  &  \ms{0.159}{0.037}  &  \ms{0.157}{0.011}  &  \ms{0.341}{0.032}  &  \ms{\textbf{0.517}}{0.032} \\
\dset{Coauthor-Physics} & \ms{0.365}{0.024}  &  \ms{\textbf{0.505}}{0.065}  &  OOM                &  \ms{0.098}{0.013}  &  \ms{0.034}{0.047}  &  \ms{0.424}{0.014}  &  \ms{\underline{0.445}}{0.026} \\
\dset{Amazon-Computers} & \ms{0.183}{0.008}  &  \ms{\underline{0.275}}{0.046}  &  \ms{0.214}{0.052}  &  \ms{0.181}{0.015}  &  \ms{0.155}{0.012}  &  \ms{0.265}{0.032}  &  \ms{\textbf{0.305}}{0.029} \\
\dset{Amazon-Photos}    & \ms{0.419}{0.126}  &  \ms{\textbf{0.461}}{0.014}  &  \ms{0.251}{0.010}  &  \ms{0.265}{0.044}  &  \ms{0.172}{0.084}  &  \ms{\underline{0.442}}{0.028}  &  \ms{0.416}{0.027} \\
\bottomrule
\end{tabular}
}
\vspace{-0.2in}
\end{table}
}
\end{savenotes}

\subsubsection{Inductive Evaluation}
\label{subsubsec:inductive_eval}
\textbf{Inductive Setting.} While we observe some promising results in favor of non-contrastive methods (namely, BGRL) in the transductive setting, we note that this setting is not entirely realistic. In practice, we often have both new nodes and edges introduced at inference time after our model is trained. For example, consider a social network upon which a model is trained at some time $t_1$ but is used for inference (for a GNN, this refers to the message-passing step) at time $t_2$, where new users and friendships have been added to the network in the interim. Then, the goal of a model run at time $t_2$ would be to predict any new links at new network state $t_3$ (although we assume there are no new nodes introduced at that step since we cannot compute the embedding of nodes without performing inference on them first). To simulate this setting, we first partition the graph into two sets of nodes: ``observed'' nodes (that we see during training) and ``unobserved nodes'' (that are only used for inference and testing). We then withhold a portion of the edges at each of the time steps $t_3, t_2, t_1$ to serve as testing-only, inference-only, and training-only edges, respectively. We describe this process in more detail in \cref{subsec:inductive_details}. %

\textbf{Results.} \cref{tab:inductive} shows that in the inductive setting, BGRL is outperformed by the contrastive ML-GCN on \textit{all} datasets. It still outperforms CCA-SSG and GBT, but it is \textit{much} less competitive in the inductive setting. We next ask: what accounts for this large difference in performance? %

\textbf{Why Does BGRL Not Work Well in the Inductive Setting?} One possible reason for the poor performance of BGRL in the inductive setting is that it is unable to correctly differentiate unseen positive from unseen negatives, i.e., it is overfitting on the training graph.  Intuitively, this could happen due to a lack of negative samples---BGRL never pushes samples away from each other. \revAdd{We show that this is indeed the case in \cref{fig:ind_dists}, where BGRL's negative link score distribution has heavy overlap with its positive link score distribution.} We can also see this behavior in \cref{fig:trans_dists} where the ML-GCN does a clearly better job of pushing positive/negative samples far apart, despite BGRL's surprising success. Naturally, improving the separation between these distributions increases the chance of a correct prediction. We investigate this hypothesis in \cref{sec:method} below \revAdd{and propose \methodname{} (\cref{fig:arch}), a novel method to help alleviate this issue}.

\vspace{-0.12in}
\section{Improving Inductive Performance in a Non-Contrastive Framework}
\label{sec:method}
\vspace{-0.1in}

\begin{figure}[t]
    \vspace{-0.2in}
    \centering
    {\includegraphics[width=0.48\textwidth]{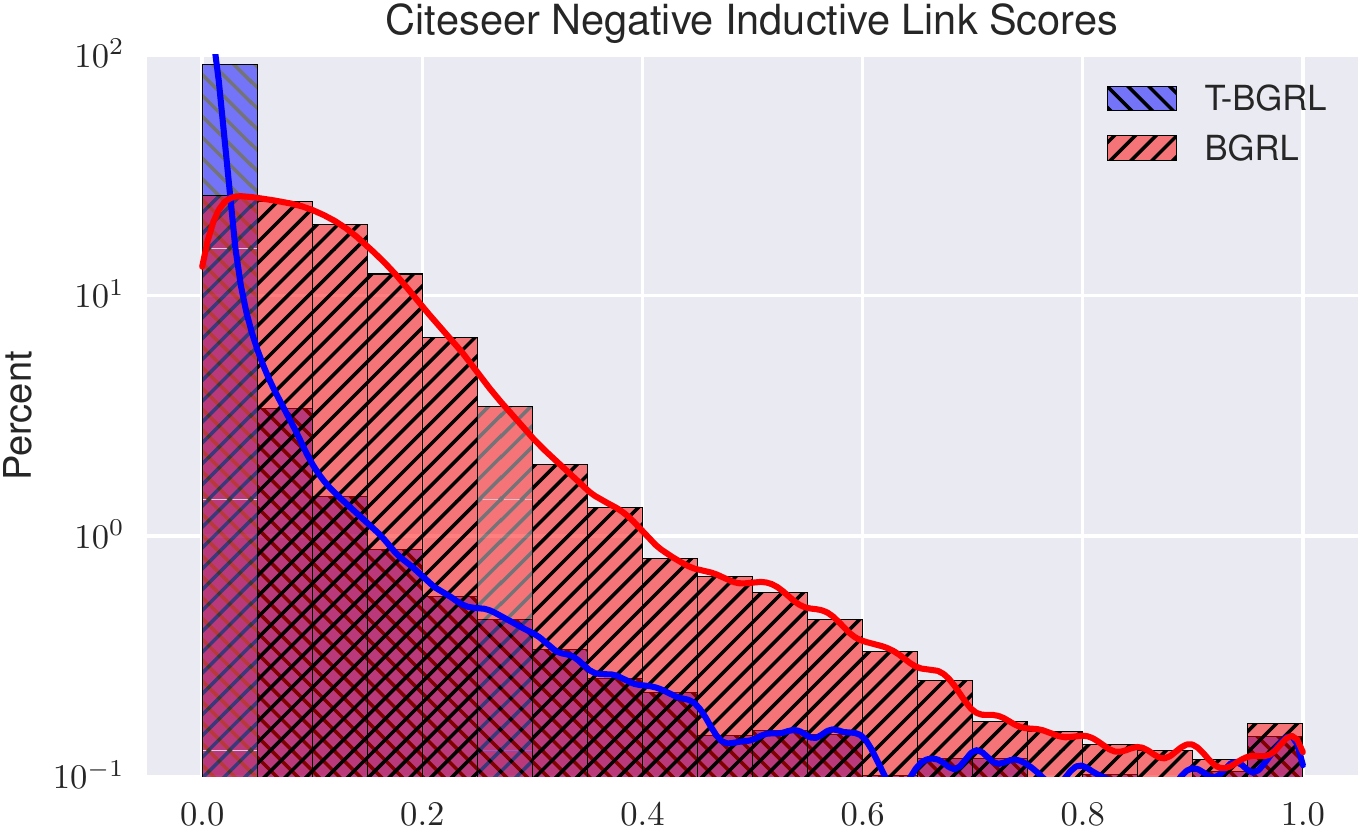}\label{fig:ind_neg_dist}}
    \hfill
    {\includegraphics[width=0.48\textwidth]{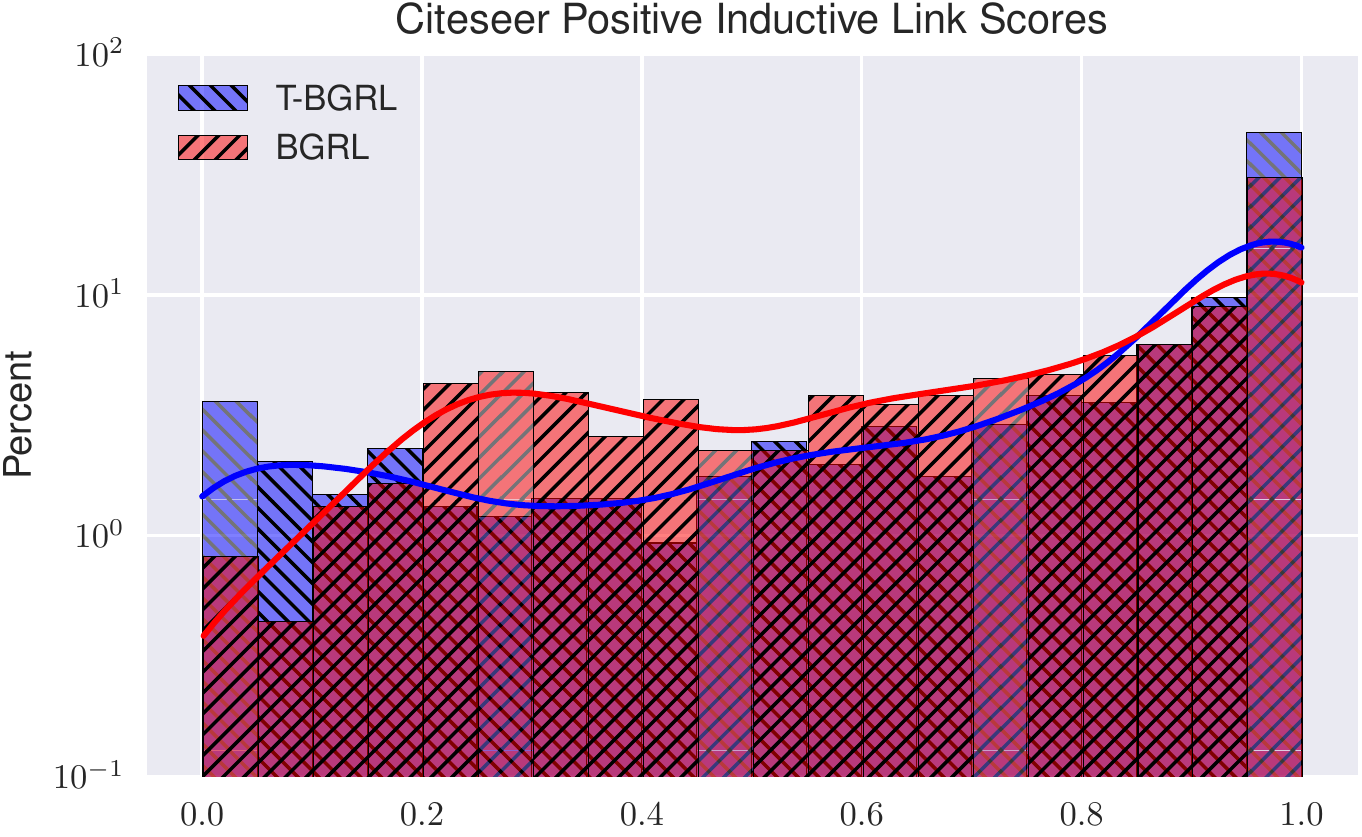}\label{fig:ind_pos_dist}}
    \caption{\revAdd{These plots show similarities between node embeddings on \dset{Citeseer}.} \textbf{Left:} distribution of similarity to \textit{non-neighbors} for \methodname{} and BGRL. \textbf{Right:} distribution of similarity to \textit{neighbors} for \methodname{} and BGRL. Note that the y-axis is on a logarithmic scale. \methodname{} clearly does a better job of ensuring that negative link representations are pushed far apart from those of positive links.}
    \label{fig:ind_dists}
\end{figure}

\begin{figure}[t]
    \centering
    \includegraphics[width=0.9\textwidth]{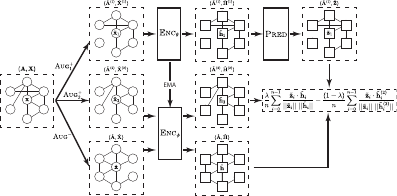}
    \caption{\methodname{} architecture diagram. The loss function is also shown in \cref{eqn:tbgrl_loss}.}
    \label{fig:arch}
    \vspace{-0.15in}
\end{figure}

In order to reduce this systematic gap in performance between ML-GCN (the best-performing contrastive model) and BGRL (the best-performing non-contrastive model), %
we observe that we need to push negative and positive node pair representations further apart. %
This way, pairs between new nodes---introduced at inference time---have a higher chance of being classified correctly.  Contrastive methods utilize negative sampling for this purpose, but we wish to avoid negative sampling owing to high computational cost. In lieu of this, we propose a simple, yet powerfully effective idea below. 

\begin{wrapfigure}[19]{R}{0.5\textwidth}
\begin{flushright}
\vspace{-0.25in}
\begin{minipage}[r]{.5\textwidth}
\begin{algorithm}[H]
    \caption{PyTorch-style pseudocode for \methodname{}}
    \label{alg:triplet}
    \definecolor{codeblue}{rgb}{0.25,0.5,0.5}
    \definecolor{codekw}{rgb}{0.85, 0.18, 0.50}
    \lstset{
      backgroundcolor=\color{white},
      basicstyle=\fontsize{7.5pt}{7.5pt}\ttfamily\selectfont,
      columns=fullflexible,
      breaklines=true,
      captionpos=b,
      commentstyle=\fontsize{7.5pt}{7.5pt}\color{codeblue},
      keywordstyle=\fontsize{7.5pt}{7.5pt}\color{codekw},
    }
\vspace{-4pt}
\begin{lstlisting}[language=python]
# Enc_o: online encoder network
# Enc_t: target encoder network
# Pred: predictor network
# lam: trade-off
# decay: EMA decay parameter
# g: input graph
# feat: node features

g1, feat1 = augment(g, feat)      # augmentation #1
g2, feat2 = augment(g, feat)      # augmentation #2
c_g, c_feat = corrupt(g, feat)  # corruption

h1 = Enc_o(g1, feat1)
h2 = Enc_t(g2, feat2)
c_z = Enc_t(c_g, c_feat)
z1 = Pred(z1)

loss = lam*cosine_similarity(z1, c_z) \
        - (1-lam)*cosine_similarity(z1, h2)
loss.backward() # backprop

# Update Enc_t with EMA
Enc_t.params = decay * Enc_t.params \
                 + (1-decay) * Enc_o.params
\end{lstlisting}
\vspace{-8pt}
\end{algorithm}
\end{minipage}
\end{flushright}
\end{wrapfigure}

\textbf{Model Intuition.} To emulate the effect of negative sampling without actually performing it, we propose Triplet-BGRL (\methodname{}). In addition to the two augmentations performed during standard non-contrastive SSL training, we add a corruption to function as a cheap negative sample. For each node, like BGRL, we minimize the distance between its representations across two augmentations. However, taking inspiration from triplet-style losses~\citep{triplet}, we also maximize the distance between the augmentation and corruption representations. %

\textbf{Model Design.}
Ideally, this model should not only perform better than BGRL in the inductive setting, but should also have the same time complexity as BGRL. In order to meet these expectations, we design efficient, linear-time corruptions (same asymptotic runtime as the augmentations). We also choose to use the online encoder $\Enc_{\phi}$ to generate embeddings for the corrupted graph so that \methodname{} does not have any additional parameters. \cref{fig:arch} illustrates the overall architecture of the proposed \methodname{}, and \cref{alg:triplet} presents PyTorch-style pseudocode. Our new proposed loss function is as follows: %
\begin{equation}%
\label{eqn:tbgrl_loss}%
\gL_{\methodname{}} = \underbrace{
    \frac{\lambda}{n} \sum^{n-1}_{i=0} \frac{\tilde{\vz}_i \cdot \check{\vh}_i}{||\tilde{\vz}_i||\ ||\check{\vh}_i||}
}_{\methodname{}\text{ Loss Term}} \underbrace{
    - \frac{(1-\lambda)}{n} \sum^{n-1}_{i=0} \frac{\tilde{\vz}_i \cdot \vh_i^{(2)}}{||\tilde{\vz}_i||\ ||\vh_i^{(2)}||}
}_{\text{BGRL Loss}}
\end{equation}%
where $\lambda$ is a hyperparameter controlling the repulsive forces between augmentation and corruption.

\textbf{Corruption Choice.} We experiment with several different corruptions methods, but limit ourselves to linear-time corruptions in order to maintain the efficiency of BGRL. We find that $\revAdd{\textsc{ShuffleFeatRandomEdge}}\maxPair = \nMaxPair$, where $\check{\mA}{\sim}\{0, 1\}^{n \times n}$ and $\check{\mX} = \textsc{ShuffleRows}(\mX)$ works the best. We describe each of the different corruptions we experimented with in \cref{subsec:more_corruptions}.

\textbf{Inductive Results.} \cref{tab:inductive} shows that \methodname{} improves inductive performance over BGRL in 5/6 datasets, with very large improvements in the \dset{Cora} and \dset{Citeseer} datasets. The only dataset where BGRL outperformed \methodname{} is the \dset{Amazon-Photos} dataset. However, this gap is much smaller (0.01 difference in Hits@50) than the improvements on the other datasets. We plot the scores output by the decoder for unseen negative pairs compared to those for unseen positive pairs in \cref{fig:ind_dists}. We can see that \methodname{} pushes apart unseen negative and positive pairs much better than BGRL.

\begin{wrapfigure}[10]{R}{0.5\textwidth}
\begin{flushright}
\vspace{-0.45in}
\begin{minipage}[r]{.5\textwidth}
\begin{table}[H]
\caption{Transductive performance of \methodname{} compared to ML-GCN and BGRL (same numbers as \cref{tab:transductive} above; full figure in \cref{tab:transductive_full}).}
\vspace{-0.08in}
\label{tab:tbgrl_transductive}
\centering
\resizebox{1\columnwidth}{!}{
\begin{tabular}{l|c|cc} 
\toprule
\textbf{Dataset}          	& \textbf{ML-GCN}          	& \textbf{BGRL}             	& \methodname{} \\
\midrule
\dset{Cora}             	& \textbf{0.815} 	& \underline{0.792} 	& \ms{0.773}{0.020}  \\
\dset{Citeseer}         	& 0.771  	& \underline{0.858} 	& \ms{\textbf{0.868}}{0.023} \\
\dset{Coauthor-Cs}      	& \textbf{0.787} 	& 0.515 	& \ms{\underline{0.555}}{0.009} \\
\dset{Coauthor-Physics} 	& \textbf{0.810}  	& 0.476 	& \ms{0.471}{0.021} \\
\dset{Amazon-Computers} 	& \underline{0.320}  	& \textbf{0.346} 	& \ms{0.315}{0.015} \\
\dset{Amazon-Photos}    	& 0.430  	& \textbf{0.562} 	& \ms{\underline{0.517}}{0.016} \\
\bottomrule
\end{tabular}
}
\end{table}
\end{minipage}
\end{flushright}
\end{wrapfigure}

\textbf{Transductive Results. } We also evaluate the performance of \methodname{} in the transductive setting to ensure that it does not significantly reduce performance when compared to BGRL.  See \cref{tab:tbgrl_transductive} on the right for the results.

\begin{wrapfigure}[16]{R}{0.5\textwidth}
\begin{flushright}
\vspace{-0.56in}
\begin{minipage}[r]{.5\textwidth}
\begin{figure}[H]
    \centering
    {\includegraphics[width=\textwidth]{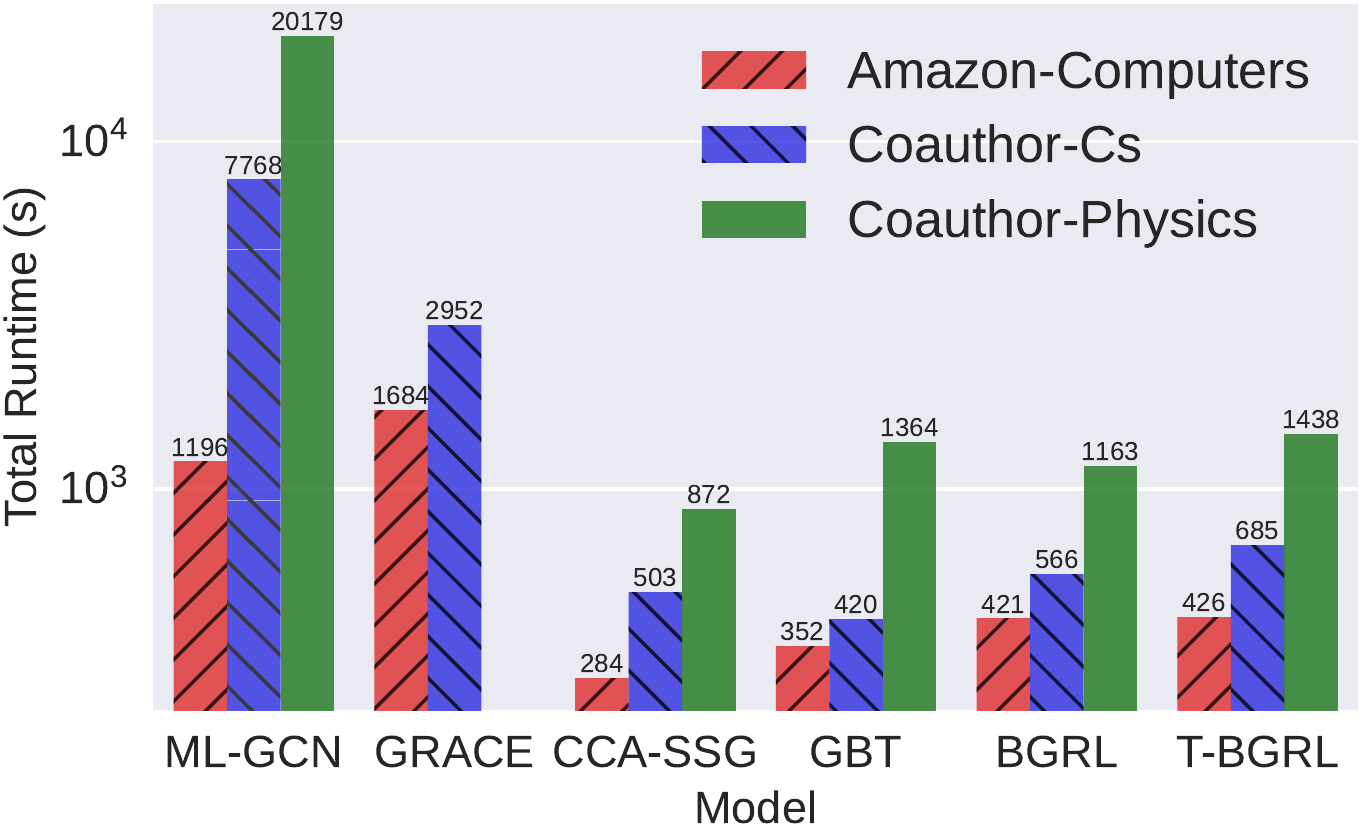}\label{fig:neg_dist}}
    \vspace{-0.25in}
    \caption{Total runtime comparison of different contrastive and non-contrastive methods. T-BGRL and BGRL have relatively similar runtimes and are significantly faster than the contrastive methods (GRACE and ML-GCN).}
    \label{fig:rel_runtime}
\end{figure}
\end{minipage}
\end{flushright}
\end{wrapfigure}

\textbf{Difference from Contrastive Methods.} %
While our method shares some similarities with contrastive methods, we believe \methodname{} is strictly non-contrastive because it does not require the $O(n^2)$ sampling from the complement of the edge index used by contrastive methods. \revAdd{This is clearly shown in \cref{fig:rel_runtime}, where T-BGRL and BGRL have similar runtimes and are much faster than GRACE and ML-GCN.} The corruption can be viewed as a ``negative'' augmentation---with the only difference being that it changes the expected label for each link. \revAdd{In fact, one of the corruptions that we consider, \textsc{SparsifyFeatSparsifyEdge}, is essentially the same as the augmentations using by BGRL (except with much higher drop probability).}  We discuss other corruptions below in \cref{subsec:more_corruptions}.

\textbf{Scalability.} We evaluate the runtime of our model on different datasets. \cref{fig:rel_runtime} shows the running times to fully train a model for different contrastive and non-contrastive methods over 5 different runs. Note that we use a fixed 10,000 epochs for GRACE, CCA-SSG, GBT, BGRL, and \methodname{}, but use early stopping on the ML-GCN with a maximum of 1,000 epochs. We find that (i) \methodname{} is comparable to BGRL in runtime owing to efficient choices of corruptions, (ii) it is about $4.3\times$ faster than GRACE on \dset{Amazon-Computers} (the largest dataset which GRACE can run on), and (ii) it is $14\times$ faster than ML-GCN. CCA-SSG is the fastest of all the methods but performs the worst. As mentioned above, we do not compare with SEAL \citep{seal} or other subgraph-based methods due to how slow they are during inference. SUREL~\citep{surel} is \textasciitilde 250\texttimes{} slower\revAdd{, and SEAL~\citep{seal} is about \textasciitilde 3900\texttimes{} slower} according to \cite{surel}. \revAdd{In conclusion, we find that \methodname{} is roughly as scalable as other non-contrastive methods, and much more scalable than existing contrastive methods.}

\vspace{-0.1in}
\section{Other Related Work}
\vspace{-0.1in}

\textbf{Link Prediction.} Link prediction is a long-standing graph machine learning task. Some traditional methods include (i) matrix~\citep{menon2011link,wang2020svd++} or tensor factorization~\citep{linkPredEvolving09,dunlavy2011temporal} methods which factor the adjacency and/or feature matrices to derive node representations which can predict links equipped with inner products, and (ii) heuristic methods which score node pairs based on neighborhood and overlap~\citep{yu2017similarity,zareie2020similarity,philip2010link}. %
Several shallow graph embedding methods~\citep{node2vec,deepWalk14} which train node embeddings by random-walk strategies have also been used for link prediction. %
In addition to the node-embedding-based GNN methods mentioned in \cref{sec:prelims}, several works \citep{surel, seal, deal} propose subgraph-based methods for this task, which aim to classify subgraphs around each candidate link.  Few works focus on scalable link prediction with distillation \citep{guo2022linkless}, decoder \citep{wang2022flashlight}, and sketching designs \citep{chamberlain2022graph}.

\textbf{Graph SSL Methods.} Most graph SSL methods can be put categorized into contrastive and non-contrastive methods. Contrastive learning has been applied to link prediction with margin-loss-based methods such as PinSAGE~\citep{pinSAGE}, and GraphSAGE~\citep{graphsage}, where negative sample representations are pushed apart from positive sample representations. GRACE~\citep{grace} uses augmentation~\citep{zhao2022graph} during this negative sampling process to further increase the performance of the model.  DGI~\citep{dgi} leverages mutual information maximization between local patch and global graph representations. Some works \citep{ju2022multi, jin2021automated} also explore using multiple contrastive pretext tasks for SSL.  Several works \citep{you2020graph, lin2022prototypical} also focus on graph-level contrastive learning, via graph-level augmentations and careful negative selection. 
 Recently, non-contrastive methods have been applied to graph representation learning. Self-GNN~\citep{selfgnn} and BGRL~\citep{bgrl} use ideas from BYOL~\citep{byol} and SimSiam~\citep{simsiam} to propose an graph framework that does not require negative sampling. We describe BGRL in depth in \cref{sec:prelims} above. Graph Barlow Twins (GBT)~\citep{gbt} is adapted from the Barlow Twins model in the image domain~\citep{barlowTwins}, and uses cross-correlation to learn node representations with a shared encoder. CCA-SSG~\citep{ccaSSG} uses ideas from Canonical Correlation Analysis (CCA)~\citep{cca} and Deep CCA~\citep{deepCCA} for their loss function. These models are somewhat similar in that it has also been shown that Barlow Twins is equivalent to Kernel CCA~\citep{balestriero2022contrastive}.

\vspace{-0.1in}
\section{Conclusion}
\vspace{-0.1in}
To our knowledge, this is the first work to study non-contrastive SSL methods and their performance on link prediction. We first evaluate several contrastive and non-contrastive graph SSL methods on link prediction tasks, and find that surprisingly, one popular non-contrastive method (BGRL) is able to perform well in the transductive setting.
We also observe that BGRL struggles in the inductive setting, and identify that it has a tendency to overfit the training graph, indicating it fails to push positive and negative node pair representations far apart from each other. Armed with these insights, we propose \methodname, a simple but effective non-contrastive strategy which works by generating extremely cheap ``negatives'' by corrupting the original inputs. \methodname sidesteps the expensive negative sampling step evident in contrastive learning, while enjoying strong performance benefits. \methodname improves on BGRL's inductive performance in 5/6 datasets while achieving similar transductive performance, making it comparable to the best contrastive baselines, but with a 14\texttimes{} speedup over the best contrastive methods.

\vspace{-0.1in}
\section*{Reproducibility Statement}
\vspace{-0.1in}
To ensure reproducibility, our source code is available 
online at \url{https://github.com/snap-research/non-contrastive-link-prediction}.
The hyperparameters and instructions for reproducing all experiments are provided in the \verb+README.md+ file.

\subsubsection*{Acknowledgments}
UCR coauthors were partly supported by the National Science Foundation under CAREER grant no. IIS 2046086 and were also sponsored by the Combat Capabilities Development Command Army Research Laboratory under Cooperative Agreement Number W911NF-13-2-0045 (ARL Cyber Security CRA). Any opinions, findings, and conclusions or recommendations expressed in this material are those of the author(s) and do not necessarily reflect the views of the funding parties.

\bibliography{refs}

\begin{thebibliography}{72}
\providecommand{\natexlab}[1]{#1}
\providecommand{\url}[1]{\texttt{#1}}
\expandafter\ifx\csname urlstyle\endcsname\relax
  \providecommand{\doi}[1]{doi: #1}\else
  \providecommand{\doi}{doi: \begingroup \urlstyle{rm}\Url}\fi

\bibitem[Acar et~al.(2009)Acar, Dunlavy, and Kolda]{linkPredEvolving09}
Evrim Acar, Daniel~M. Dunlavy, and Tamara~G. Kolda.
\newblock Link prediction on evolving data using matrix and tensor
  factorizations.
\newblock In \emph{2009 IEEE International Conference on Data Mining
  Workshops}, pp.\  262--269, 2009.
\newblock \doi{10.1109/ICDMW.2009.54}.

\bibitem[Andrew et~al.(2013)Andrew, Arora, Bilmes, and Livescu]{deepCCA}
Galen Andrew, Raman Arora, Jeff Bilmes, and Karen Livescu.
\newblock Deep canonical correlation analysis.
\newblock In \emph{International conference on machine learning}, pp.\
  1247--1255. PMLR, 2013.

\bibitem[Balestriero \& LeCun(2022)Balestriero and
  LeCun]{balestriero2022contrastive}
Randall Balestriero and Yann LeCun.
\newblock Contrastive and non-contrastive self-supervised learning recover
  global and local spectral embedding methods.
\newblock \emph{arXiv preprint arXiv:2205.11508}, 2022.

\bibitem[Berg et~al.(2017)Berg, Kipf, and Welling]{berg2017graph}
Rianne van~den Berg, Thomas~N Kipf, and Max Welling.
\newblock Graph convolutional matrix completion.
\newblock \emph{arXiv preprint arXiv:1706.02263}, 2017.

\bibitem[Bielak et~al.(2022)Bielak, Kajdanowicz, and Chawla]{gbt}
Piotr Bielak, Tomasz Kajdanowicz, and Nitesh~V. Chawla.
\newblock Graph barlow twins: A self-supervised representation learning
  framework for graphs.
\newblock \emph{Knowledge-Based Systems}, pp.\  109631, 2022.
\newblock ISSN 0950-7051.
\newblock \doi{https://doi.org/10.1016/j.knosys.2022.109631}.
\newblock URL
  \url{https://www.sciencedirect.com/science/article/pii/S095070512200822X}.

\bibitem[Biewald(2020)]{wandb}
Lukas Biewald.
\newblock Experiment tracking with weights and biases, 2020.
\newblock URL \url{https://www.wandb.com/}.
\newblock Software available from wandb.com.

\bibitem[Cai et~al.(2020)Cai, Li, Wang, and Ji]{lgnlp}
Lei Cai, Jundong Li, Jie Wang, and Shuiwang Ji.
\newblock Line graph neural networks for link prediction, 2020.
\newblock URL \url{https://arxiv.org/abs/2010.10046}.

\bibitem[Chamberlain et~al.(2022)Chamberlain, Shirobokov, Rossi, Frasca,
  Markovich, Hammerla, Bronstein, and Hansmire]{chamberlain2022graph}
Benjamin~Paul Chamberlain, Sergey Shirobokov, Emanuele Rossi, Fabrizio Frasca,
  Thomas Markovich, Nils Hammerla, Michael~M Bronstein, and Max Hansmire.
\newblock Graph neural networks for link prediction with subgraph sketching.
\newblock \emph{arXiv preprint arXiv:2209.15486}, 2022.

\bibitem[Chen et~al.(2019)Chen, Ye, Zuo, Zheng, and Ong]{chen2019graph}
Chi Chen, Weike Ye, Yunxing Zuo, Chen Zheng, and Shyue~Ping Ong.
\newblock Graph networks as a universal machine learning framework for
  molecules and crystals.
\newblock \emph{Chemistry of Materials}, 31\penalty0 (9):\penalty0 3564--3572,
  2019.

\bibitem[Chen et~al.(2018)Chen, Yin, Wang, Wang, Nguyen, and Li]{pme}
Hongxu Chen, Hongzhi Yin, Weiqing Wang, Hao Wang, Quoc Viet~Hung Nguyen, and
  Xue Li.
\newblock Pme: Projected metric embedding on heterogeneous networks for link
  prediction.
\newblock In \emph{Proceedings of the 24th ACM SIGKDD International Conference
  on Knowledge Discovery \& Data Mining}, KDD '18, pp.\  1177–1186, New York,
  NY, USA, 2018. Association for Computing Machinery.
\newblock ISBN 9781450355520.
\newblock \doi{10.1145/3219819.3219986}.
\newblock URL \url{https://doi.org/10.1145/3219819.3219986}.

\bibitem[Chen et~al.(2020)Chen, Kornblith, Norouzi, and Hinton]{simclr}
Ting Chen, Simon Kornblith, Mohammad Norouzi, and Geoffrey~E. Hinton.
\newblock A simple framework for contrastive learning of visual
  representations.
\newblock In \emph{ICML}, Proceedings of Machine Learning Research, pp.\
  1597--1607, 2020.

\bibitem[Chen \& He(2021)Chen and He]{simsiam}
Xinlei Chen and Kaiming He.
\newblock Exploring simple siamese representation learning.
\newblock In \emph{Proceedings of the IEEE/CVF Conference on Computer Vision
  and Pattern Recognition}, pp.\  15750--15758, 2021.

\bibitem[Cremonesi et~al.(2010)Cremonesi, Koren, and Turrin]{recsysPerf}
Paolo Cremonesi, Yehuda Koren, and Roberto Turrin.
\newblock Performance of recommender algorithms on top-n recommendation tasks.
\newblock In \emph{Proceedings of the Fourth ACM Conference on Recommender
  Systems}, RecSys '10, pp.\  39–46, New York, NY, USA, 2010. Association for
  Computing Machinery.
\newblock ISBN 9781605589060.
\newblock \doi{10.1145/1864708.1864721}.
\newblock URL \url{https://doi.org/10.1145/1864708.1864721}.

\bibitem[Derrow-Pinion et~al.(2021)Derrow-Pinion, She, Wong, Lange, Hester,
  Perez, Nunkesser, Lee, Guo, Wiltshire, et~al.]{derrow2021eta}
Austin Derrow-Pinion, Jennifer She, David Wong, Oliver Lange, Todd Hester, Luis
  Perez, Marc Nunkesser, Seongjae Lee, Xueying Guo, Brett Wiltshire, et~al.
\newblock Eta prediction with graph neural networks in google maps.
\newblock In \emph{Proceedings of the 30th ACM International Conference on
  Information \& Knowledge Management}, pp.\  3767--3776, 2021.

\bibitem[Dunlavy et~al.(2011)Dunlavy, Kolda, and Acar]{dunlavy2011temporal}
Daniel~M Dunlavy, Tamara~G Kolda, and Evrim Acar.
\newblock Temporal link prediction using matrix and tensor factorizations.
\newblock \emph{ACM Transactions on Knowledge Discovery from Data (TKDD)},
  5\penalty0 (2):\penalty0 1--27, 2011.

\bibitem[Fan \& Huang(2019)Fan and Huang]{fan2019labeled}
Shuangfei Fan and Bert Huang.
\newblock Labeled graph generative adversarial networks.
\newblock \emph{arXiv preprint arXiv:1906.03220}, 2019.

\bibitem[Fan et~al.(2022)Fan, Liu, Jin, Zhao, Tang, and Li]{fan2022graph}
Wenqi Fan, Xiaorui Liu, Wei Jin, Xiangyu Zhao, Jiliang Tang, and Qing Li.
\newblock Graph trend filtering networks for recommendation.
\newblock In \emph{Proceedings of the 45th International ACM SIGIR Conference
  on Research and Development in Information Retrieval}, pp.\  112--121, 2022.

\bibitem[Grill et~al.(2020)Grill, Strub, Altch{\'e}, Tallec, Richemond,
  Buchatskaya, Doersch, Avila~Pires, Guo, Gheshlaghi~Azar, et~al.]{byol}
Jean-Bastien Grill, Florian Strub, Florent Altch{\'e}, Corentin Tallec, Pierre
  Richemond, Elena Buchatskaya, Carl Doersch, Bernardo Avila~Pires, Zhaohan
  Guo, Mohammad Gheshlaghi~Azar, et~al.
\newblock Bootstrap your own latent-a new approach to self-supervised learning.
\newblock \emph{Advances in neural information processing systems},
  33:\penalty0 21271--21284, 2020.

\bibitem[Grover \& Leskovec(2016)Grover and Leskovec]{node2vec}
Aditya Grover and Jure Leskovec.
\newblock node2vec: Scalable feature learning for networks.
\newblock In \emph{KDD}, pp.\  855--864. {ACM}, 2016.

\bibitem[Guo et~al.(2021)Guo, Zhang, Yu, Herr, Wiest, Jiang, and
  Chawla]{guo2021few}
Zhichun Guo, Chuxu Zhang, Wenhao Yu, John Herr, Olaf Wiest, Meng Jiang, and
  Nitesh~V Chawla.
\newblock Few-shot graph learning for molecular property prediction.
\newblock In \emph{Proceedings of the Web Conference 2021}, pp.\  2559--2567,
  2021.

\bibitem[Guo et~al.(2022{\natexlab{a}})Guo, Nan, Tian, Wiest, Zhang, and
  Chawla]{guo2022graph}
Zhichun Guo, Bozhao Nan, Yijun Tian, Olaf Wiest, Chuxu Zhang, and Nitesh~V
  Chawla.
\newblock Graph-based molecular representation learning.
\newblock \emph{arXiv preprint arXiv:2207.04869}, 2022{\natexlab{a}}.

\bibitem[Guo et~al.(2022{\natexlab{b}})Guo, Shiao, Zhang, Liu, Chawla, Shah,
  and Zhao]{guo2022linkless}
Zhichun Guo, William Shiao, Shichang Zhang, Yozen Liu, Nitesh Chawla, Neil
  Shah, and Tong Zhao.
\newblock Linkless link prediction via relational distillation.
\newblock \emph{arXiv preprint arXiv:2210.05801}, 2022{\natexlab{b}}.

\bibitem[Hamilton et~al.(2017)Hamilton, Ying, and Leskovec]{graphsage}
William~L. Hamilton, Zhitao Ying, and Jure Leskovec.
\newblock Inductive representation learning on large graphs.
\newblock In \emph{NIPS}, pp.\  1024--1034, 2017.

\bibitem[Hao et~al.(2020)Hao, Cao, Fang, Xie, and Wang]{deal}
Yu~Hao, Xin Cao, Yixiang Fang, Xike Xie, and Sibo Wang.
\newblock Inductive link prediction for nodes having only attribute
  information.
\newblock In \emph{Proceedings of the Twenty-Ninth International Joint
  Conference on Artificial Intelligence}. International Joint Conferences on
  Artificial Intelligence Organization, jul 2020.
\newblock \doi{10.24963/ijcai.2020/168}.
\newblock URL \url{https://doi.org/10.24963%2Fijcai.2020%2F168}.

\bibitem[Hassani \& Ahmadi(2020)Hassani and Ahmadi]{mvgrl}
Kaveh Hassani and Amir Hosein~Khas Ahmadi.
\newblock Contrastive multi-view representation learning on graphs.
\newblock In \emph{ICML}, volume 119 of \emph{Proceedings of Machine Learning
  Research}, pp.\  4116--4126. {PMLR}, 2020.

\bibitem[He et~al.(2020)He, Deng, Wang, Li, Zhang, and Wang]{he2020lightgcn}
Xiangnan He, Kuan Deng, Xiang Wang, Yan Li, Yongdong Zhang, and Meng Wang.
\newblock Lightgcn: Simplifying and powering graph convolution network for
  recommendation.
\newblock In \emph{Proceedings of the 43rd International ACM SIGIR conference
  on research and development in Information Retrieval}, pp.\  639--648, 2020.

\bibitem[Hoffer \& Ailon(2014)Hoffer and Ailon]{triplet}
Elad Hoffer and Nir Ailon.
\newblock Deep metric learning using triplet network, 2014.
\newblock URL \url{https://arxiv.org/abs/1412.6622}.

\bibitem[Hotelling(1992)]{cca}
Harold Hotelling.
\newblock Relations between two sets of variates.
\newblock In \emph{Breakthroughs in statistics}, pp.\  162--190. Springer,
  1992.

\bibitem[Hu et~al.(2020)Hu, Fey, Zitnik, Dong, Ren, Liu, Catasta, and
  Leskovec]{ogb}
Weihua Hu, Matthias Fey, Marinka Zitnik, Yuxiao Dong, Hongyu Ren, Bowen Liu,
  Michele Catasta, and Jure Leskovec.
\newblock Open graph benchmark: Datasets for machine learning on graphs.
\newblock \emph{arXiv preprint arXiv:2005.00687}, 2020.

\bibitem[Hubert et~al.(2022)Hubert, Monnin, Brun, and
  Monticolo]{newStratsRecommendation}
Nicolas Hubert, Pierre Monnin, Armelle Brun, and Davy Monticolo.
\newblock {New Strategies for Learning Knowledge Graph Embeddings: the
  Recommendation Case}.
\newblock In \emph{{EKAW 2022 - 23rd International Conference on Knowledge
  Engineering and Knowledge Management}}, Bolzano, Italy, September 2022.
\newblock URL \url{https://hal.inria.fr/hal-03722881}.

\bibitem[Jin et~al.(2020)Jin, Derr, Liu, Wang, Wang, Liu, and
  Tang]{jin2020self}
Wei Jin, Tyler Derr, Haochen Liu, Yiqi Wang, Suhang Wang, Zitao Liu, and
  Jiliang Tang.
\newblock Self-supervised learning on graphs: Deep insights and new direction.
\newblock \emph{arXiv preprint arXiv:2006.10141}, 2020.

\bibitem[Jin et~al.(2021)Jin, Liu, Zhao, Ma, Shah, and Tang]{jin2021automated}
Wei Jin, Xiaorui Liu, Xiangyu Zhao, Yao Ma, Neil Shah, and Jiliang Tang.
\newblock Automated self-supervised learning for graphs.
\newblock \emph{arXiv preprint arXiv:2106.05470}, 2021.

\bibitem[Ju et~al.(2022)Ju, Zhao, Wen, Yu, Shah, Ye, and Zhang]{ju2022multi}
Mingxuan Ju, Tong Zhao, Qianlong Wen, Wenhao Yu, Neil Shah, Yanfang Ye, and
  Chuxu Zhang.
\newblock Multi-task self-supervised graph neural networks enable stronger task
  generalization.
\newblock \emph{arXiv preprint arXiv:2210.02016}, 2022.

\bibitem[Kefato \& Girdzijauskas(2021)Kefato and Girdzijauskas]{selfgnn}
Zekarias~T. Kefato and Sarunas Girdzijauskas.
\newblock Self-supervised graph neural networks without explicit negative
  sampling.
\newblock \emph{CoRR}, abs/2103.14958, 2021.
\newblock URL \url{https://arxiv.org/abs/2103.14958}.

\bibitem[Kipf \& Welling(2017)Kipf and Welling]{gcn}
Thomas~N. Kipf and Max Welling.
\newblock Semi-supervised classification with graph convolutional networks.
\newblock In \emph{ICLR}, 2017.

\bibitem[Lin et~al.(2022)Lin, Liu, Zhou, Hu, Wang, Zhao, Zheng, Lin, Xing, and
  Liang]{lin2022prototypical}
Shuai Lin, Chen Liu, Pan Zhou, Zi-Yuan Hu, Shuojia Wang, Ruihui Zhao, Yefeng
  Zheng, Liang Lin, Eric Xing, and Xiaodan Liang.
\newblock Prototypical graph contrastive learning.
\newblock \emph{IEEE Transactions on Neural Networks and Learning Systems},
  2022.

\bibitem[Lin et~al.(2015)Lin, Liu, Sun, Liu, and Zhu]{learningEntityKGC}
Yankai Lin, Zhiyuan Liu, Maosong Sun, Yang Liu, and Xuan Zhu.
\newblock Learning entity and relation embeddings for knowledge graph
  completion.
\newblock In \emph{Proceedings of the Twenty-Ninth AAAI Conference on
  Artificial Intelligence}, AAAI'15, pp.\  2181–2187. AAAI Press, 2015.
\newblock ISBN 0262511290.

\bibitem[Liu et~al.(2022)Liu, Zhao, Xu, Luo, and Jiang]{liu2022graph}
Gang Liu, Tong Zhao, Jiaxin Xu, Tengfei Luo, and Meng Jiang.
\newblock Graph rationalization with environment-based augmentations.
\newblock In \emph{Proceedings of the 28th ACM SIGKDD Conference on Knowledge
  Discovery and Data Mining}, pp.\  1069--1078, 2022.

\bibitem[Menon \& Elkan(2011)Menon and Elkan]{menon2011link}
Aditya~Krishna Menon and Charles Elkan.
\newblock Link prediction via matrix factorization.
\newblock In \emph{Joint european conference on machine learning and knowledge
  discovery in databases}, pp.\  437--452. Springer, 2011.

\bibitem[Oord et~al.(2018)Oord, Li, and Vinyals]{oord2018representation}
Aaron van~den Oord, Yazhe Li, and Oriol Vinyals.
\newblock Representation learning with contrastive predictive coding.
\newblock \emph{arXiv preprint arXiv:1807.03748}, 2018.

\bibitem[Perozzi et~al.(2014)Perozzi, Al-Rfou, and Skiena]{deepWalk14}
Bryan Perozzi, Rami Al-Rfou, and Steven Skiena.
\newblock {DeepWalk}.
\newblock In \emph{Proceedings of the 20th {ACM} {SIGKDD} international
  conference on Knowledge discovery and data mining}. {ACM}, aug 2014.
\newblock \doi{10.1145/2623330.2623732}.
\newblock URL \url{https://doi.org/10.1145%2F2623330.2623732}.

\bibitem[Philip et~al.(2010)Philip, Han, and Faloutsos]{philip2010link}
S~Yu Philip, Jiawei Han, and Christos Faloutsos.
\newblock \emph{Link mining: Models, algorithms, and applications}.
\newblock Springer, 2010.

\bibitem[Rendle et~al.(2020)Rendle, Krichene, Zhang, and
  Anderson]{rendle2020neural}
Steffen Rendle, Walid Krichene, Li~Zhang, and John Anderson.
\newblock Neural collaborative filtering vs. matrix factorization revisited.
\newblock In \emph{Fourteenth ACM conference on recommender systems}, pp.\
  240--248, 2020.

\bibitem[Sankar et~al.(2021)Sankar, Liu, Yu, and Shah]{sankar2021graph}
Aravind Sankar, Yozen Liu, Jun Yu, and Neil Shah.
\newblock Graph neural networks for friend ranking in large-scale social
  platforms.
\newblock In \emph{Proceedings of the Web Conference 2021}, pp.\  2535--2546,
  2021.

\bibitem[Sen et~al.(2008)Sen, Namata, Bilgic, Getoor, Galligher, and
  Eliassi-Rad]{coraCiteseer}
Prithviraj Sen, Galileo Namata, Mustafa Bilgic, Lise Getoor, Brian Galligher,
  and Tina Eliassi-Rad.
\newblock Collective classification in network data.
\newblock \emph{AI magazine}, 2008.

\bibitem[Shchur et~al.(2018)Shchur, Mumme, Bojchevski, and
  Gunnemann]{amazonComputersPhotos}
Oleksandr Shchur, Maximilian Mumme, Aleksandar Bojchevski, and Stephan
  Gunnemann.
\newblock Pitfalls of graph neural network evaluation.
\newblock \emph{arXiv preprint arXiv:1811.05868}, 2018.

\bibitem[Shiao \& Papalexakis(2021)Shiao and
  Papalexakis]{shiao2021adversarially}
William Shiao and Evangelos~E Papalexakis.
\newblock Adversarially generating rank-constrained graphs.
\newblock In \emph{2021 IEEE 8th International Conference on Data Science and
  Advanced Analytics (DSAA)}, pp.\  1--8. IEEE, 2021.

\bibitem[Tang et~al.(2020)Tang, Liu, Shah, Shi, Mitra, and
  Wang]{tang2020knowing}
Xianfeng Tang, Yozen Liu, Neil Shah, Xiaolin Shi, Prasenjit Mitra, and Suhang
  Wang.
\newblock Knowing your fate: Friendship, action and temporal explanations for
  user engagement prediction on social apps.
\newblock In \emph{Proceedings of the 26th ACM SIGKDD international conference
  on knowledge discovery \& data mining}, pp.\  2269--2279, 2020.

\bibitem[Tang et~al.(2022)Tang, Liu, He, Wang, and Shah]{tang2022friend}
Xianfeng Tang, Yozen Liu, Xinran He, Suhang Wang, and Neil Shah.
\newblock Friend story ranking with edge-contextual local graph convolutions.
\newblock In \emph{Proceedings of the Fifteenth ACM International Conference on
  Web Search and Data Mining}, pp.\  1007--1015, 2022.

\bibitem[Thakoor et~al.(2022)Thakoor, Tallec, Azar, Azabou, Dyer, Munos,
  Velickovic, and Valko]{bgrl}
Shantanu Thakoor, Corentin Tallec, Mohammad~Gheshlaghi Azar, Mehdi Azabou,
  Eva~L. Dyer, R{\'{e}}mi Munos, Petar Velickovic, and Michal Valko.
\newblock Large-scale representation learning on graphs via bootstrapping.
\newblock In \emph{The Tenth International Conference on Learning
  Representations, {ICLR} 2022, Virtual Event, April 25-29, 2022}.
  OpenReview.net, 2022.
\newblock URL \url{https://openreview.net/forum?id=0UXT6PpRpW}.

\bibitem[Tian et~al.(2021)Tian, Chen, and Ganguli]{understandingSSL}
Yuandong Tian, Xinlei Chen, and Surya Ganguli.
\newblock Understanding self-supervised learning dynamics without contrastive
  pairs, 2021.
\newblock URL \url{https://arxiv.org/abs/2102.06810}.

\bibitem[Veličković et~al.(2018)Veličković, Fedus, Hamilton, Liò, Bengio,
  and Hjelm]{dgi}
Petar Veličković, William Fedus, William~L. Hamilton, Pietro Liò, Yoshua
  Bengio, and R~Devon Hjelm.
\newblock Deep graph infomax, 2018.
\newblock URL \url{https://arxiv.org/abs/1809.10341}.

\bibitem[Wang et~al.(2020)Wang, Sun, and Li]{wang2020svd++}
Shijie Wang, Guiling Sun, and Yangyang Li.
\newblock Svd++ recommendation algorithm based on backtracking.
\newblock \emph{Information}, 11\penalty0 (7):\penalty0 369, 2020.

\bibitem[Wang et~al.(2022)Wang, Hooi, Liu, Zhao, Guo, and
  Shah]{wang2022flashlight}
Yiwei Wang, Bryan Hooi, Yozen Liu, Tong Zhao, Zhichun Guo, and Neil Shah.
\newblock Flashlight: Scalable link prediction with effective decoders.
\newblock \emph{arXiv preprint arXiv:2209.10100}, 2022.

\bibitem[Wang et~al.(2021)Wang, Zhou, Hong, Zou, and Su]{wang2021pairwise}
Zhitao Wang, Yong Zhou, Litao Hong, Yuanhang Zou, and Hanjing Su.
\newblock Pairwise learning for neural link prediction.
\newblock \emph{arXiv preprint arXiv:2112.02936}, 2021.

\bibitem[Wen \& Li(2022)Wen and Li]{wen2022mechanism}
Zixin Wen and Yuanzhi Li.
\newblock The mechanism of prediction head in non-contrastive self-supervised
  learning.
\newblock \emph{arXiv preprint arXiv:2205.06226}, 2022.

\bibitem[Yang et~al.(2020)Yang, Ding, Zhou, Yang, Zhou, and
  Tang]{understanding_neg_sampling}
Zhen Yang, Ming Ding, Chang Zhou, Hongxia Yang, Jingren Zhou, and Jie Tang.
\newblock Understanding negative sampling in graph representation learning,
  2020.
\newblock URL \url{https://arxiv.org/abs/2005.09863}.

\bibitem[Yin et~al.(2022)Yin, Zhang, Wang, Wang, and Li]{surel}
Haoteng Yin, Muhan Zhang, Yanbang Wang, Jianguo Wang, and Pan Li.
\newblock Algorithm and system co-design for efficient subgraph-based graph
  representation learning.
\newblock \emph{Proceedings of the VLDB Endowment}, 15\penalty0 (11):\penalty0
  2788--2796, 2022.

\bibitem[Ying et~al.(2018{\natexlab{a}})Ying, He, Chen, Eksombatchai, Hamilton,
  and Leskovec]{pinSAGE}
Rex Ying, Ruining He, Kaifeng Chen, Pong Eksombatchai, William~L. Hamilton, and
  Jure Leskovec.
\newblock Graph convolutional neural networks for web-scale recommender
  systems.
\newblock In \emph{Proceedings of the 24th {ACM} {SIGKDD} International
  Conference on Knowledge Discovery \& Data Mining}. {ACM}, jul
  2018{\natexlab{a}}.
\newblock \doi{10.1145/3219819.3219890}.
\newblock URL \url{https://doi.org/10.1145%2F3219819.3219890}.

\bibitem[Ying et~al.(2018{\natexlab{b}})Ying, He, Chen, Eksombatchai, Hamilton,
  and Leskovec]{ying2018graph}
Rex Ying, Ruining He, Kaifeng Chen, Pong Eksombatchai, William~L Hamilton, and
  Jure Leskovec.
\newblock Graph convolutional neural networks for web-scale recommender
  systems.
\newblock In \emph{Proceedings of the 24th ACM SIGKDD international conference
  on knowledge discovery \& data mining}, pp.\  974--983, 2018{\natexlab{b}}.

\bibitem[You et~al.(2018)You, Ying, Ren, Hamilton, and Leskovec]{graphrnn}
Jiaxuan You, Rex Ying, Xiang Ren, William Hamilton, and Jure Leskovec.
\newblock Graphrnn: Generating realistic graphs with deep auto-regressive
  models.
\newblock In \emph{International conference on machine learning}, pp.\
  5708--5717. PMLR, 2018.

\bibitem[You et~al.(2020)You, Chen, Sui, Chen, Wang, and Shen]{you2020graph}
Yuning You, Tianlong Chen, Yongduo Sui, Ting Chen, Zhangyang Wang, and Yang
  Shen.
\newblock Graph contrastive learning with augmentations.
\newblock \emph{Advances in Neural Information Processing Systems},
  33:\penalty0 5812--5823, 2020.

\bibitem[Yu et~al.(2017)Yu, Zhao, An, and Lin]{yu2017similarity}
Chuanming Yu, Xiaoli Zhao, Lu~An, and Xia Lin.
\newblock Similarity-based link prediction in social networks: A path and node
  combined approach.
\newblock \emph{Journal of Information Science}, 43\penalty0 (5):\penalty0
  683--695, 2017.

\bibitem[Zareie \& Sakellariou(2020)Zareie and
  Sakellariou]{zareie2020similarity}
Ahmad Zareie and Rizos Sakellariou.
\newblock Similarity-based link prediction in social networks using latent
  relationships between the users.
\newblock \emph{Scientific Reports}, 10\penalty0 (1):\penalty0 1--11, 2020.

\bibitem[Zbontar et~al.(2021)Zbontar, Jing, Misra, LeCun, and
  Deny]{barlowTwins}
Jure Zbontar, Li~Jing, Ishan Misra, Yann LeCun, and Stéphane Deny.
\newblock Barlow twins: Self-supervised learning via redundancy reduction,
  2021.
\newblock URL \url{https://arxiv.org/abs/2103.03230}.

\bibitem[Zhang et~al.(2021)Zhang, Wu, Yan, Wipf, and Yu]{ccaSSG}
Hengrui Zhang, Qitian Wu, Junchi Yan, David Wipf, and Philip~S Yu.
\newblock From canonical correlation analysis to self-supervised graph neural
  networks.
\newblock \emph{Advances in Neural Information Processing Systems},
  34:\penalty0 76--89, 2021.

\bibitem[Zhang \& Chen(2018)Zhang and Chen]{seal}
Muhan Zhang and Yixin Chen.
\newblock Link prediction based on graph neural networks.
\newblock In \emph{Advances in Neural Information Processing Systems}, pp.\
  5165--5175, 2018.

\bibitem[Zhang \& Chen(2019)Zhang and Chen]{zhang2019inductive}
Muhan Zhang and Yixin Chen.
\newblock Inductive matrix completion based on graph neural networks.
\newblock \emph{arXiv preprint arXiv:1904.12058}, 2019.

\bibitem[Zhang et~al.(2020)Zhang, Cui, and Zhu]{zhang2020deep}
Ziwei Zhang, Peng Cui, and Wenwu Zhu.
\newblock Deep learning on graphs: A survey.
\newblock \emph{IEEE Transactions on Knowledge and Data Engineering}, 2020.

\bibitem[Zhao et~al.(2022{\natexlab{a}})Zhao, Jin, Liu, Wang, Liu, Günneman,
  Shah, and Jiang]{zhao2022graph}
Tong Zhao, Wei Jin, Yozen Liu, Yingheng Wang, Gang Liu, Stephan Günneman, Neil
  Shah, and Meng Jiang.
\newblock Graph data augmentation for graph machine learning: A survey.
\newblock \emph{arXiv preprint arXiv:2202.08871}, 2022{\natexlab{a}}.

\bibitem[Zhao et~al.(2022{\natexlab{b}})Zhao, Liu, Wang, Yu, and
  Jiang]{zhao2022learning}
Tong Zhao, Gang Liu, Daheng Wang, Wenhao Yu, and Meng Jiang.
\newblock Learning from counterfactual links for link prediction.
\newblock In \emph{International Conference on Machine Learning}, pp.\
  26911--26926. PMLR, 2022{\natexlab{b}}.

\bibitem[Zhu et~al.(2020)Zhu, Xu, Yu, Liu, Wu, and Wang]{grace}
Yanqiao Zhu, Yichen Xu, Feng Yu, Qiang Liu, Shu Wu, and Liang Wang.
\newblock Deep graph contrastive representation learning.
\newblock \emph{arXiv preprint arXiv:2006.04131}, 2020.

\end{thebibliography}
\bibliographystyle{iclr2023_conference}

\appendix
\section{Appendix}

\subsection{Dataset Statistics}
\label{subsec:dset_stats}
\begin{table}[H]
\centering
\caption{Statistics for the datasets used in our work.}
\label{tab:dset_statistic}
\begin{tabular}{l||ccc} 
\toprule
\textbf{Dataset}    & \textbf{Nodes}    & \textbf{Edges}   & \textbf{Features}    \\ \midrule
\dset{Cora}    &2,708 & 5,278 &1,433 \\
\dset{Citeseer} &3,327 &4,552 &3,703 \\
\dset{Coauthor-Cs} &18,333 &163,788 &6,805 \\   
\dset{Coauthor-Physics} &34,493 &495,924 &8,415 \\      
\dset{Amazon-Computers} &13,752 &491,722 &767 \\
\dset{Amazon-Photos} &7,650 &238,162 &745 \\

\bottomrule
\end{tabular}
\end{table}

\subsection{Machine Details}
We run all of our experiments on either NVIDIA P100 or V100 GPUs. We use machines with 12 virtual CPU cores and 24 GB of RAM for the majority of our experiments. We exclusively use V100s for our timing experiments. We ran our experiments on Google Cloud Platform.

\subsection{Transductive Setting Details}
We use a 85/5/10 split for training/validation/testing data---following \citet{seal,lgnlp}.

\subsection{Inductive Setting Details}
\label{subsec:inductive_details}
The inductive setting represents a more realistic setting than the transductive setting. For example, consider a social network upon which a model is trained at some time $t_1$ but is used for inference (for a GNN, this refers to the message-passing step) at time $t_2$, where new users and friendships have been added to the network in the interim. Then, the goal of a model run at time $t_2$ would be to predict any new links at new network state $t_3$ (although we assume there are no new nodes introduced at that step since we cannot compute the embedding of nodes without performing inference on them first). To simulate this setting, we first perform the following steps:

\begin{enumerate}
    \item We withhold a portion of the edges (and same number of disconnected node pairs) to use as testing-only edges.
    \item We partition the graph into two sets of nodes: ``observed'' nodes (that we see during training) and ``unobserved nodes'' (that can only be seen during testing).
    \item We mask out some edges to use as testing-only edges.
    \item We mask out some edges to use as inference-only edges.
    \item We mask out some edges to use as validation-only edges.
    \item We mask out some edges to use as training-only edges.
\end{enumerate}
As the test edges are sampled before the node split, there will be three kinds of them after the splitting. Specifically: edges within observed nodes, edges between observed nodes and unobserved nodes, and edges within unobserved nodes. For the ease of data preparation, we use the same percentages for the test edge splitting, unobserved node splitting, and validation edge splitting. Specifically, we mask out 30\% of the edges (at each of the above stages) on the small datasets (\dset{Cora} and \dset{Citeseer}), and 10\% on all the other datasets. We use a 30\% split on the small datasets to ensure that we have a sufficient number of edges for testing and validation purposes.

\subsection{Experimental Setup}
To ensure that we fairly evaluate each model, we run a Bayesian hyperparameter sweep for 25 runs across each model-dataset combination with the target metric being the validation Hits@50. Each run is the result of the mean averaged over 5 runs (retraining both the encoder and decoder). We used the Weights and Biases~\citep{wandb} Bayesian optimizer for our experiments. We provide a sample configuration file to reproduce our sweeps, as well as the exact parameters used for the top \methodname{} runs shown in our tables.

We used the reference GRACE implementation and BGRL implementation, but modified them for link prediction instead of node classification. We based our E2E-GCN off of the OGB~\citep{ogb} implementation. We re-implemented CCA-SSG and GBT. The code for all of our implementations and modifications can be found in the link in our paper above.

\subsection{Full Results}
\cref{tab:transductive_full} shows the results of all the methods (including \methodname{}) on transductive setting.

\begin{table}[H]
\centering
\caption{Full transductive performance table (combination of \cref{tab:transductive,tab:tbgrl_transductive}).}
\label{tab:transductive_full}
\resizebox{\linewidth}{!}{%
\begin{tabular}{l||c||cc|ccc|c} 
\toprule
&\multicolumn{1}{c||}{\textbf{End-To-End}} & \multicolumn{2}{c|}{\textbf{Contrastive}} & \multicolumn{4}{c}{\textbf{Non-Contrastive}} \\
\midrule
\textbf{Dataset} & \textbf{E2E-GCN} & \textbf{ML-GCN} & \textbf{GRACE} & \textbf{CCA-SSG} & \textbf{GBT} & \textbf{BGRL} & \textbf{T-BGRL} \\ 
\midrule
\dset{Cora} &\ms{\textbf{0.816}}{0.013} &\ms{\underline{0.815}}{0.002} &\ms{0.686}{0.056} &\ms{0.348}{0.091} &\ms{0.460}{0.149} &\ms{0.792}{0.015} &\ms{0.773}{0.020} \\
\dset{Citeseer} &\ms{0.822}{0.017} &\ms{0.771}{0.020} &\ms{0.707}{0.068} &\ms{0.249}{0.168} &\ms{0.472}{0.196} &\ms{\underline{0.858}}{0.020} &\ms{\textbf{0.868}}{0.023} \\
\dset{Amazon-Photos} &\ms{0.642}{0.029} &\ms{0.430}{0.032} &\ms{0.486}{0.025} &\ms{0.369}{0.013} &\ms{0.434}{0.038} &\ms{0.562}{0.013} &\ms{0.517}{0.016} \\
\dset{Amazon-Computers} &\ms{0.426}{0.036} &\ms{0.320}{0.060} &\ms{0.240}{0.027} &\ms{0.201}{0.032} &\ms{0.258}{0.008} &\ms{0.346}{0.018} &\ms{0.315}{0.015} \\
\dset{Coauthor-Cs} &\ms{0.762}{0.010} &\ms{0.787}{0.011} &\ms{0.456}{0.066} &\ms{0.229}{0.018} &\ms{0.298}{0.033} &\ms{0.515}{0.016} &\ms{0.555}{0.009} \\
\dset{Coauthor-Physics} &\ms{0.798}{0.018} &\ms{0.810}{0.003} &OOM &\ms{0.157}{0.009} &\ms{0.187}{0.011} &\ms{0.476}{0.015} &\ms{0.471}{0.021} \\
\bottomrule
\end{tabular}
}
\end{table}

\subsection{Corruptions}
\label{subsec:more_corruptions}

In this work, we experiment with the following corruptions:
\begin{enumerate}
    \item \textsc{RandomFeatRandomEdge}: Randomly generate an adjacency matrix $\tilde{\mA}$ and $\tilde{\mX}$ with the same sizes as $\mA$ and $\mX$, respectively. \revAdd{Note that $\tilde{\mA}$ and $\mA$ also have the same number of non-zero entries, i.e., the same number of edges.}
    \item \textsc{ShuffleFeatRandomEdge}: Randomly shuffle the rows of $\mX$, and generate a random $\tilde{\mA}$ with the same size as $\mA$. \revAdd{Note that $\tilde{\mA}$ and $\mA$ also have the same number of non-zero entries, i.e., the same number of edges.}
    \item \textsc{SparsifyFeatSparsifyEdge}: Mask out a large percentage (we chose 95\%) of the entries in $\mX$ and $\mA$.
\end{enumerate}

Of these corruptions, we find that \textsc{RandomFeatRandomEdge} works the best across our experiments.

\subsection{AUC-ROC Results}
\label{subsec:auc_res}

Here we include the area under the ROC curve for each of the different models under both the inductive and transductive settings. Note that we perform early stopping on the validation Hits@50 when training the link prediction model, not on the validation AUC-ROC.

\begin{table}[!t]
\caption{Area under the ROC curve for the methods in the transductive setting.}
\label{tab:transductive_auc}
\resizebox{\linewidth}{!}{%
\begin{tabular}{l||c||cc|cccc}
\toprule
&\multicolumn{1}{c||}{\textbf{End-To-End}} & \multicolumn{2}{c|}{\textbf{Contrastive}} & \multicolumn{4}{c}{\textbf{Non-Contrastive}} \\
\midrule
\textbf{Dataset}		& \textbf{E2E-GCN}				& \textbf{ML-GCN}				& \textbf{GRACE}	& \textbf{CCA-SSG}	& \textbf{GBT} & \textbf{BGRL} & \textbf{T-BGRL} \\ \midrule
\dset{Cora}				& \ms{\textbf{0.911}}{0.004}	& \ms{0.893}{0.007}				& \ms{0.883}{0.020}	& \ms{0.647}{0.076}	& \ms{0.736}{0.109} & \ms{\textbf{0.911}}{0.008} & \ms{\underline{0.910}}{0.005} \\
\dset{Citeseer}			& \ms{0.922}{0.006}				& \ms{0.891}{0.006}				& \ms{0.863}{0.042}	& \ms{0.661}{0.050}	& \ms{0.755}{0.120} & \ms{\underline{0.934}}{0.009} & \ms{\textbf{0.953}}{0.003} \\
\dset{Coauthor-Cs}		& \ms{\underline{0.964}}{0.005}	& \ms{\textbf{0.966}}{0.001}	& \ms{0.961}{0.003}	& \ms{0.758}{0.047}	& \ms{0.894}{0.017} & \ms{0.959}{0.002} & \ms{0.956}{0.002} \\
\dset{Coauthor-Physics}	& \ms{\underline{0.978}}{0.001}	& \ms{\textbf{0.986}}{0.000}	& OOM				& \ms{0.821}{0.051}	& \ms{0.834}{0.084} & \ms{0.961}{0.002} & \ms{0.963}{0.001} \\
\dset{Amazon-Computers}	& \ms{\textbf{0.985}}{0.001}	& \ms{\underline{0.983}}{0.001}	& \ms{0.951}{0.011}	& \ms{0.907}{0.025}	& \ms{0.946}{0.007} & \ms{0.969}{0.002} & \ms{0.976}{0.001} \\
\dset{Amazon-Photos}	& \ms{\textbf{0.989}}{0.000}	& \ms{\underline{0.983}}{0.002}	& \ms{0.981}{0.001}	& \ms{0.939}{0.008}	& \ms{0.956}{0.015} & \ms{0.980}{0.000} & \ms{0.982}{0.000} \\
\bottomrule
\end{tabular}
}
\end{table}
\begin{table}[!h]
\centering
\caption{AUC-ROC of various methods in the inductive setting. See \cref{subsubsec:inductive_eval} for an explanation of our inductive setting.}
\label{tab:inductive_auc}
\small
\resizebox{\columnwidth}{!}{
\begin{tabular}{l||c||cc|cccc}
\toprule
 &\multicolumn{1}{c||}{\textbf{End-To-End}} & \multicolumn{2}{c|}{\textbf{Contrastive}} & \multicolumn{4}{c}{\textbf{Non-Contrastive}} \\
 \midrule
 Dataset                  & \textbf{E2E-GCN}   & \textbf{ML-GCN}     & \textbf{GRACE}      & \textbf{GBT}        & \textbf{CCA-SSG}    & \textbf{BGRL}       &  \textbf{T-BGRL} \\

\midrule
\midrule
\multicolumn{8}{c}{Overall}\\ \midrule
\dset{Cora} & \ms{0.788}{0.015} & \ms{0.842}{0.008} & \ms{\underline{0.858}}{0.012} & \ms{0.704}{0.032} & \ms{0.595}{0.035} & \ms{0.814}{0.022} & \ms{\textbf{0.920}}{0.008} \\
\dset{Citeseer} & \ms{0.810}{0.016} & \ms{0.873}{0.004} & \ms{0.886}{0.010} & \ms{0.691}{0.007} & \ms{0.621}{0.070} & \ms{\underline{0.891}}{0.006} & \ms{\textbf{0.954}}{0.003} \\
\dset{Coauthor-Cs} & \ms{0.881}{0.040} & \ms{0.956}{0.001} & \ms{0.944}{0.001} & \ms{0.875}{0.036} & \ms{0.831}{0.068} & \ms{\textbf{0.968}}{0.001} & \ms{\underline{0.958}}{0.001} \\
\dset{Coauthor-Physics} & \ms{0.957}{0.004} & \ms{\textbf{0.976}}{0.001} & OOM  & \ms{0.818}{0.092} & \ms{0.614}{0.050} & \ms{\underline{0.974}}{0.001} & \ms{\textbf{0.976}}{0.001} \\
\dset{Amazon-Computers} & \ms{0.974}{0.009} & \ms{\underline{0.981}}{0.001} & \ms{0.972}{0.012} & \ms{0.919}{0.023} & \ms{0.910}{0.031} & \ms{0.980}{0.002} & \ms{\textbf{0.982}}{0.002} \\
\dset{Amazon-Photos} & \ms{0.976}{0.003} & \ms{\underline{0.982}}{0.001} & \ms{0.977}{0.002} & \ms{0.962}{0.011} & \ms{0.885}{0.057} & \ms{\textbf{0.984}}{0.000} & \ms{0.981}{0.001} \\

\midrule
\midrule
\multicolumn{8}{c}{Performance on Observed-Observed Node Edges}\\ \midrule
\dset{Cora} & \ms{0.827}{0.011} & \ms{0.834}{0.010} & \ms{\underline{0.883}}{0.010} & \ms{0.714}{0.034} & \ms{0.584}{0.047} & \ms{0.800}{0.025} & \ms{\textbf{0.929}}{0.005} \\
\dset{Citeseer} & \ms{0.792}{0.014} & \ms{0.840}{0.013} & \ms{\underline{0.905}}{0.012} & \ms{0.705}{0.019} & \ms{0.635}{0.078} & \ms{0.875}{0.006} & \ms{\textbf{0.956}}{0.005} \\
\dset{Coauthor-Cs} & \ms{0.886}{0.037} & \ms{0.951}{0.001} & \ms{0.947}{0.002} & \ms{0.874}{0.037} & \ms{0.828}{0.070} & \ms{\textbf{0.967}}{0.001} & \ms{\underline{0.955}}{0.002} \\
\dset{Coauthor-Physics} & \ms{0.959}{0.004} & \ms{\textbf{0.976}}{0.001} & OOM & \ms{0.819}{0.091} & \ms{0.615}{0.049} & \ms{0.974}{0.001} & \ms{\underline{0.975}}{0.001} \\
\dset{Amazon-Computers} & \ms{0.974}{0.010} & \ms{\textbf{0.981}}{0.001} & \ms{0.971}{0.012} & \ms{0.918}{0.024} & \ms{0.910}{0.030} & \ms{\underline{0.979}}{0.002} & \ms{\textbf{0.981}}{0.002} \\
\dset{Amazon-Photos} & \ms{0.976}{0.003} & \ms{\underline{0.981}}{0.001} & \ms{0.977}{0.002} & \ms{0.962}{0.011} & \ms{0.885}{0.055} & \ms{\textbf{0.983}}{0.000} & \ms{\underline{0.981}}{0.001} \\

\midrule
\midrule
\multicolumn{8}{c}{Performance on Observed-Unobserved Node Edges}\\ \midrule
\dset{Cora} & \ms{0.741}{0.022} & \ms{\underline{0.844}}{0.010} & \ms{0.840}{0.017} & \ms{0.696}{0.030} & \ms{0.602}{0.024} & \ms{0.818}{0.023} & \ms{\textbf{0.912}}{0.010} \\
\dset{Citeseer} & \ms{0.841}{0.019} & \ms{0.901}{0.005} & \ms{0.877}{0.012} & \ms{0.687}{0.016} & \ms{0.610}{0.069} & \ms{\underline{0.904}}{0.006} & \ms{\textbf{0.955}}{0.004} \\
\dset{Coauthor-Cs} & \ms{0.877}{0.045} & \ms{\underline{0.964}}{0.001} & \ms{0.940}{0.001} & \ms{0.876}{0.036} & \ms{0.836}{0.067} & \ms{\textbf{0.969}}{0.001} & \ms{\underline{0.964}}{0.001} \\
\dset{Coauthor-Physics} & \ms{0.953}{0.004} & \ms{\underline{0.975}}{0.001} & OOM & \ms{0.817}{0.093} & \ms{0.612}{0.052} & \ms{\underline{0.975}}{0.001} & \ms{\textbf{0.976}}{0.000} \\
\dset{Amazon-Computers} & \ms{0.974}{0.009} & \ms{\underline{0.981}}{0.001} & \ms{0.973}{0.011} & \ms{0.921}{0.022} & \ms{0.909}{0.032} & \ms{0.980}{0.002} & \ms{\textbf{0.982}}{0.002} \\
\dset{Amazon-Photos} & \ms{0.977}{0.003} & \ms{\underline{0.983}}{0.001} & \ms{0.977}{0.002} & \ms{0.963}{0.012} & \ms{0.884}{0.059} & \ms{\textbf{0.986}}{0.000} & \ms{0.981}{0.002} \\

\midrule
\midrule
\multicolumn{8}{c}{Performance on Unobserved-Unobserved Node Edges}\\ \midrule
\dset{Cora} & \ms{0.571}{0.043} & \ms{\underline{0.879}}{0.016} & \ms{0.810}{0.019} & \ms{0.693}{0.039} & \ms{0.626}{0.040} & \ms{0.866}{0.024} & \ms{\textbf{0.911}}{0.011} \\
\dset{Citeseer} & \ms{0.852}{0.047} & \ms{\underline{0.917}}{0.021} & \ms{0.827}{0.029} & \ms{0.637}{0.052} & \ms{0.599}{0.062} & \ms{0.916}{0.012} & \ms{\textbf{0.941}}{0.011} \\
\dset{Coauthor-Cs} & \ms{0.850}{0.059} & \ms{0.964}{0.001} & \ms{0.928}{0.004} & \ms{0.877}{0.034} & \ms{0.839}{0.061} & \ms{\textbf{0.967}}{0.002} & \ms{\underline{0.966}}{0.001} \\
\dset{Coauthor-Physics} & \ms{0.949}{0.006} & \ms{\underline{0.978}}{0.001} & OOM & \ms{0.818}{0.091} & \ms{0.613}{0.056} & \ms{\underline{0.978}}{0.001} & \ms{\textbf{0.981}}{0.001} \\
\dset{Amazon-Computers} & \ms{0.970}{0.010} & \ms{\textbf{0.979}}{0.001} & \ms{0.969}{0.011} & \ms{0.914}{0.022} & \ms{0.899}{0.035} & \ms{\underline{0.977}}{0.002} & \ms{\textbf{0.979}}{0.003} \\
\dset{Amazon-Photos} & \ms{0.978}{0.004} & \ms{\underline{0.982}}{0.002} & \ms{0.977}{0.002} & \ms{0.965}{0.011} & \ms{0.886}{0.063} & \ms{\textbf{0.986}}{0.001} & \ms{\underline{0.982}}{0.003} \\

\bottomrule
\end{tabular}
}
\end{table}

\subsection{Why Does BGRL Not Collapse?}
\label{subsec:why_no_collapse}

\revAdd{The loss function for BGRL (see \cref{eqn:bgrl_loss}) is 0 when $h_i^{(2)}$ = 0 or $\tilde{\vz}_i$ = 0. While theoretically possible, this is clearly undesirable behavior since this does not result in useful embeddings. We refer to this case as the model collapsing.} It is not fully understood why non-contrastive models do not collapse, but there have been several reasons proposed in the image domain with both theoretical and empirical grounding. \citet{simsiam} showed that the SimSiam architecture requires both the predictor and the stop gradient. This has also been shown to be true for BGRL. %
\citet{understandingSSL} claim that the eigenspace of predictor weights will align with the correlation matrix of the online network under the assumption of a one-layer linear encoder and a one-layer linear predictor. %
\citet{wen2022mechanism} \revAdd{looked} at the case of a two-layer non-linear encoder with output normalization and \revAdd{found} that the predictor is often only useful during the learning process, and often converges to the identity function. We did not observe this behavior on BGRL---the predictor is usually significantly different from that of the identity function.%

\subsection{How Does BGRL Pull Representations Closer Together?}
\label{subsec:pulling_repr_together}

\revAdd{Here we clarify the intuition behind BGRL pulling similar points together. To simplify this analysis, we assume that the predictor is the identity function, which \citet{wen2022mechanism} found is true in the image representation learning setting. Although we have not observed this in the graph setting, this assumption greatly simplifies our analysis and we argue it is sufficient for understanding why BGRL works.

Suppose we have three nodes: an anchor node $u$, a neighbor $v$, and a non-neighbor $w$. That is, we have $(u, v) \in \gE$, $(u, w) \not \in \gE$, and $(v, w) \not \in \gE$. Let $\vu, \vv, \vw$ be the embeddings for $u, v, w$, respectively (e.g. $\vu = \textsc{Enc}(u)$).

Assuming homophily between the nodes, we have $\vu \cdot \vv < \vu \cdot \vw$. For ease of visualization, let us project the points in a 2D space. Then, we have the following:}

\begin{figure}[!h]
\centering
\includegraphics[width=0.4\linewidth]{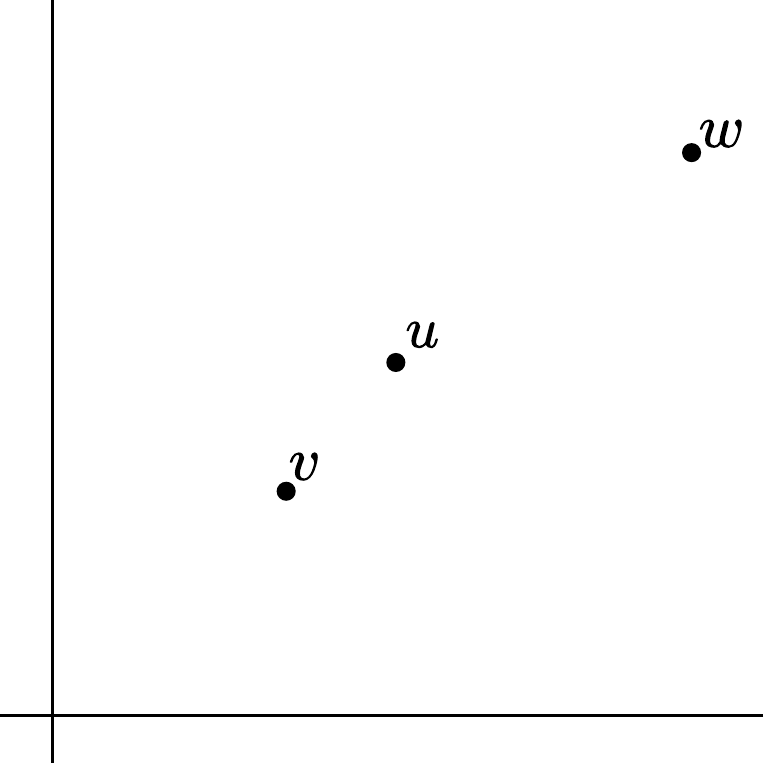}
\end{figure}
\FloatBarrier

\revAdd{We then apply the two augmentations on $u$, producing $\tilde{u}_1 = \textsc{Aug}_1 (u)$ and $\tilde{u}_2 = \textsc{Aug}_2 (u)$. For the sake of simplicity, let us assume that we perform edge dropping and feature dropping with the same probability $p$ (in practice, they may be different from each other). We represent the space of possible values for $\tilde{u}_1 $ and $\tilde{u}_2 $ as a circle with radius $r$ centered at $u$, where $r$ is controlled by $p$.}

\begin{figure}[!h]
\centering
\includegraphics[width=0.5\linewidth]{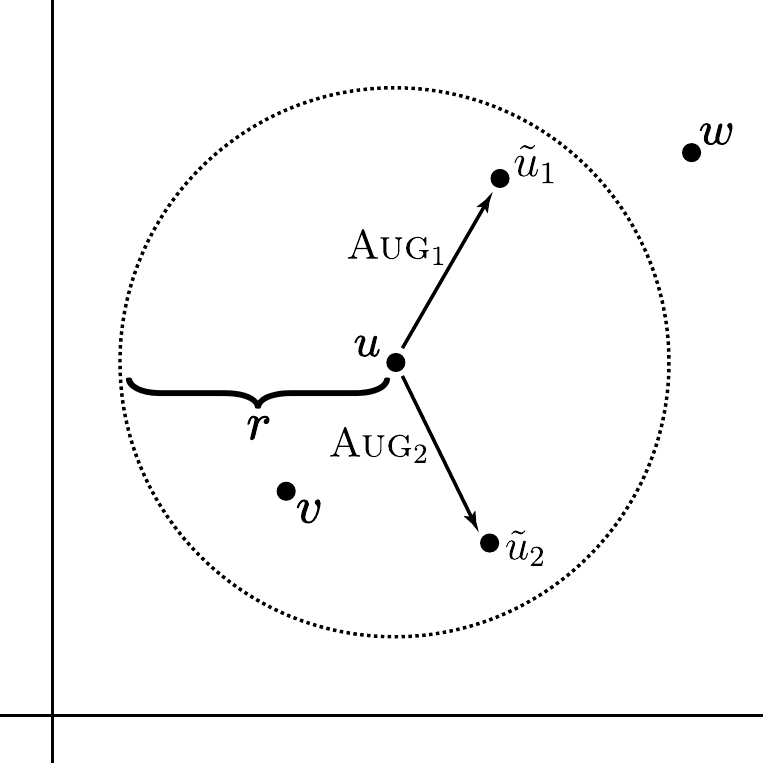}
\end{figure}
\FloatBarrier

\revAdd{The BGRL loss is stated in \cref{eqn:bgrl_loss} above, but we rewrite it relative to our anchor $u$ and with our assumption about the predictor:}%
\begin{equation}
\label{eqn:bgrl_local_loss}
    \gL_u = - \frac{\tilde{\vu}_1 \cdot \tilde{\vu}_2}{||\tilde{\vu}_1||\ ||\tilde{\vu}_2 ||}
\end{equation}

\revAdd{Minimizing this loss pushes $\tilde{\vu}_1$ and $\tilde{\vu}_2$ closer. Let us denote the encoder after one round of optimization as $\Enc'$. Then:}%
\begin{equation}
    \E \left[ || \Enc'(\textsc{Aug} (u)) - \Enc'( \textsc{Aug}(u)) || \right] < \E \left[ || \Enc( \textsc{Aug}(u)) - \Enc( \textsc{Aug}(u)) || \right]   
\end{equation}%
\revAdd{Note that $\vv$ in this example lies within the space of possible augmentations - that is, $v \in \gA$, where $\gA$ is the set of all possible values of $\textsc{Aug} (u)$. This means, as we repeat this process, we implicitly push $\vu$ and $\vv$ closer together - leading to distributions like those shown in \cref{fig:trans_dists}}.

\subsection{Additional Plots}
\label{subsec:additonal_plots}

\begin{figure}[H]
    \centering
    {\includegraphics[width=0.49\textwidth]{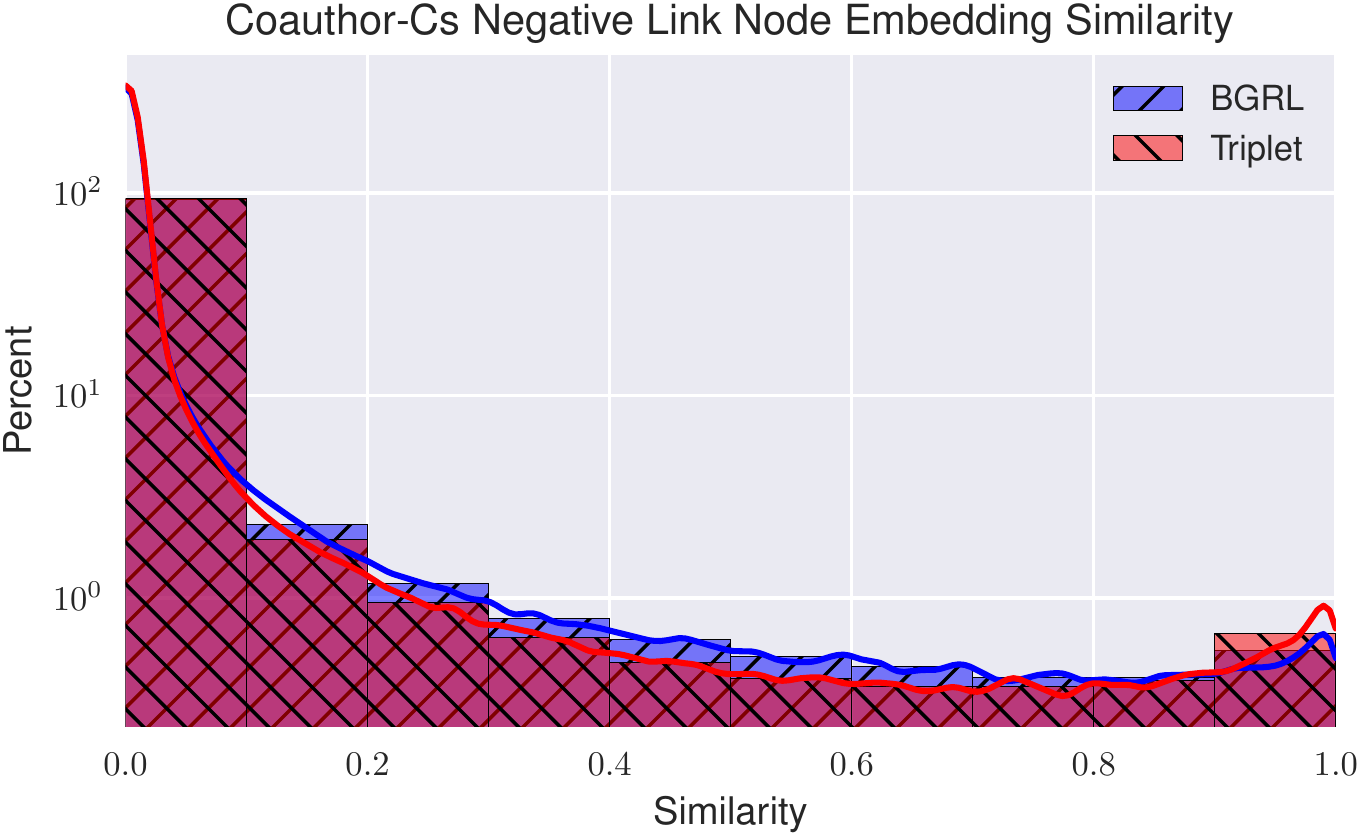}\label{fig:coauthor-cs_ind_neg_dist}}
    {\includegraphics[width=0.49\textwidth]{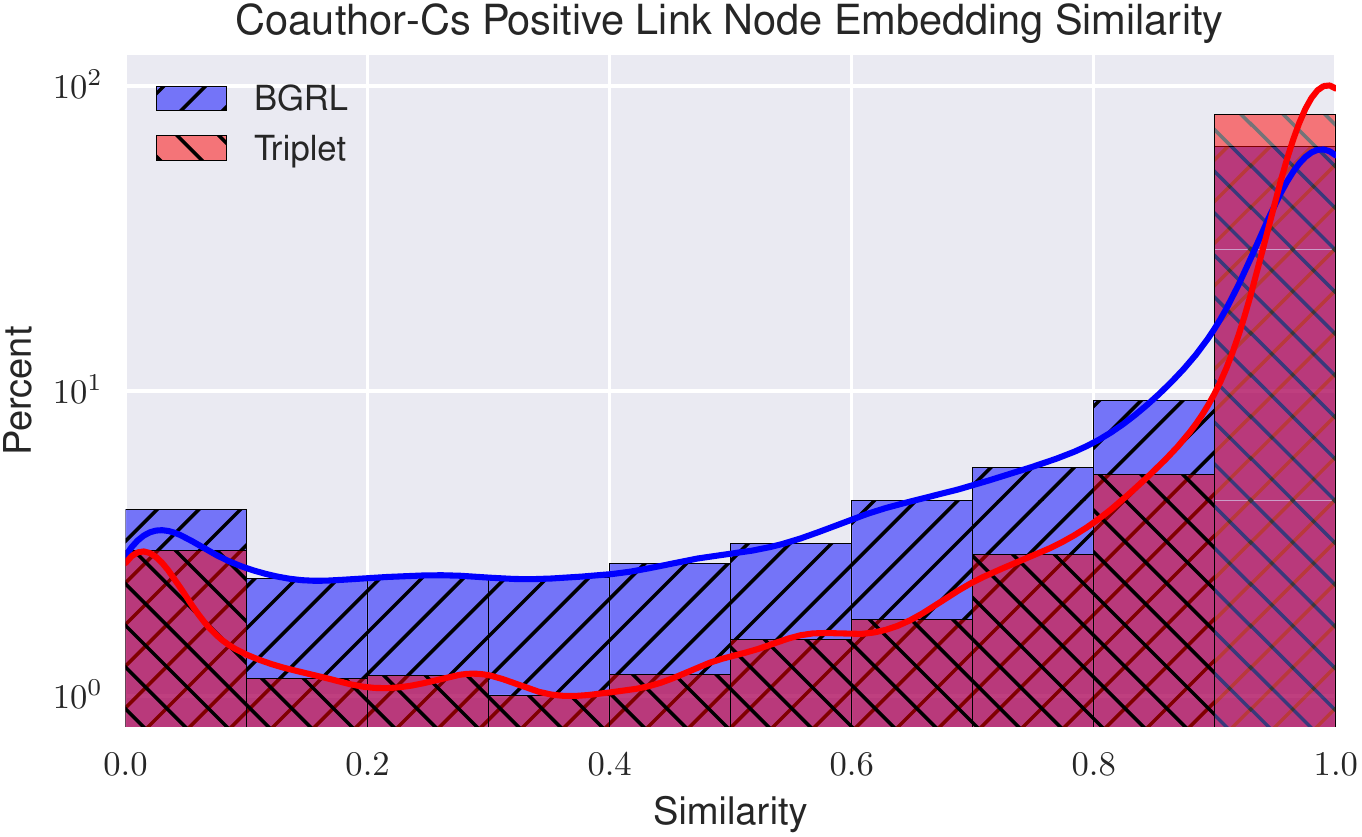}\label{fig:coauthor-cs_ind_pos_dist}}
    \caption{\revAdd{These plots show similarities between node embeddings on \dset{Coauthor-Cs}. \textbf{Left:} distribution of similarity to \textit{non-neighbors} for \methodname{} and BGRL (closer to 0 is better). \textbf{Right:} distribution of similarity to \textit{neighbors} for \methodname{} and BGRL (closer to 1 is better). Note that the y-axis is on a logarithmic scale. \methodname{} clearly does a better job of ensuring that negative link representations are pushed far apart from those of positive links.}}
    \label{fig:coauthor-cs_ind_dists}
\end{figure}

\begin{figure}[H]
    \centering
    {\includegraphics[width=0.49\textwidth]{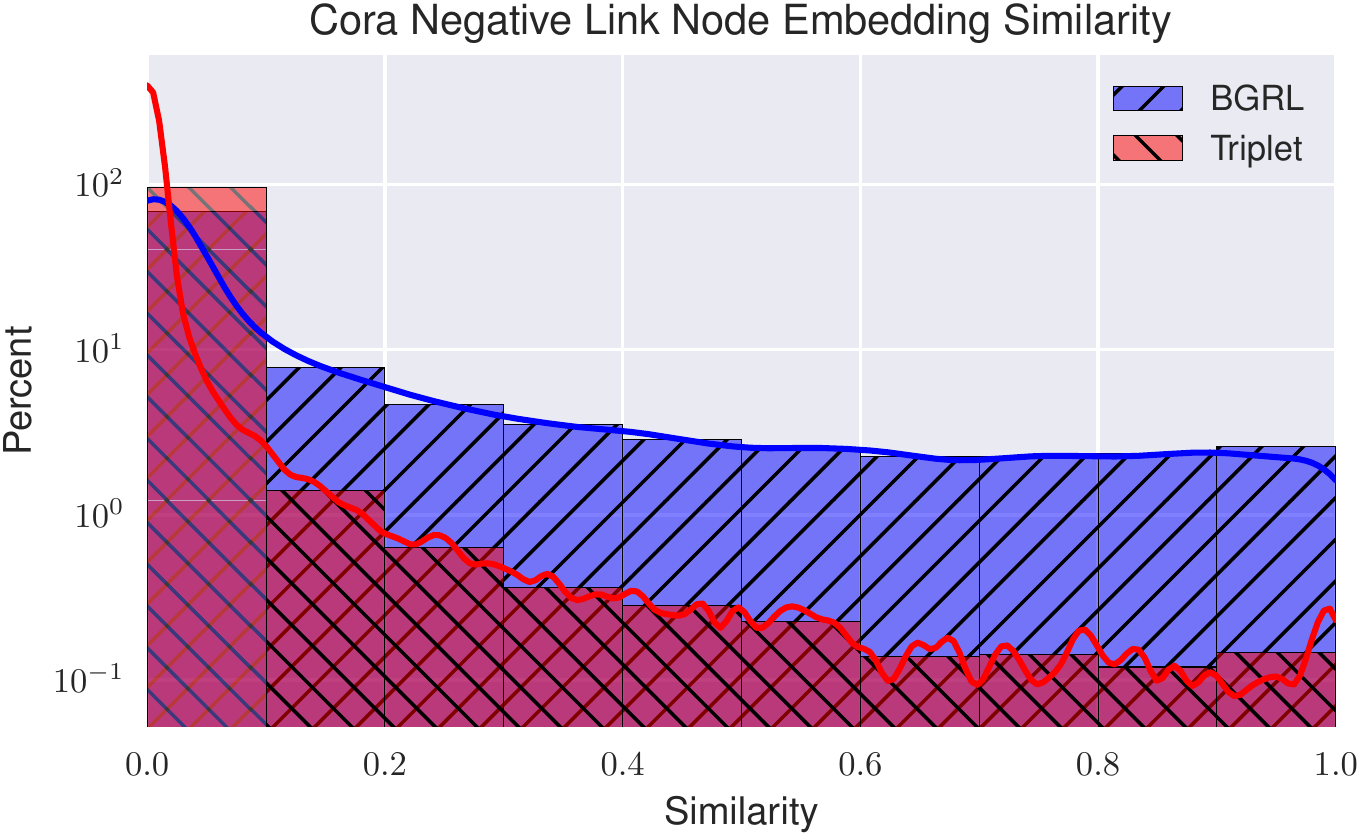}\label{fig:cora_ind_neg_dist}}
    {\includegraphics[width=0.49\textwidth]{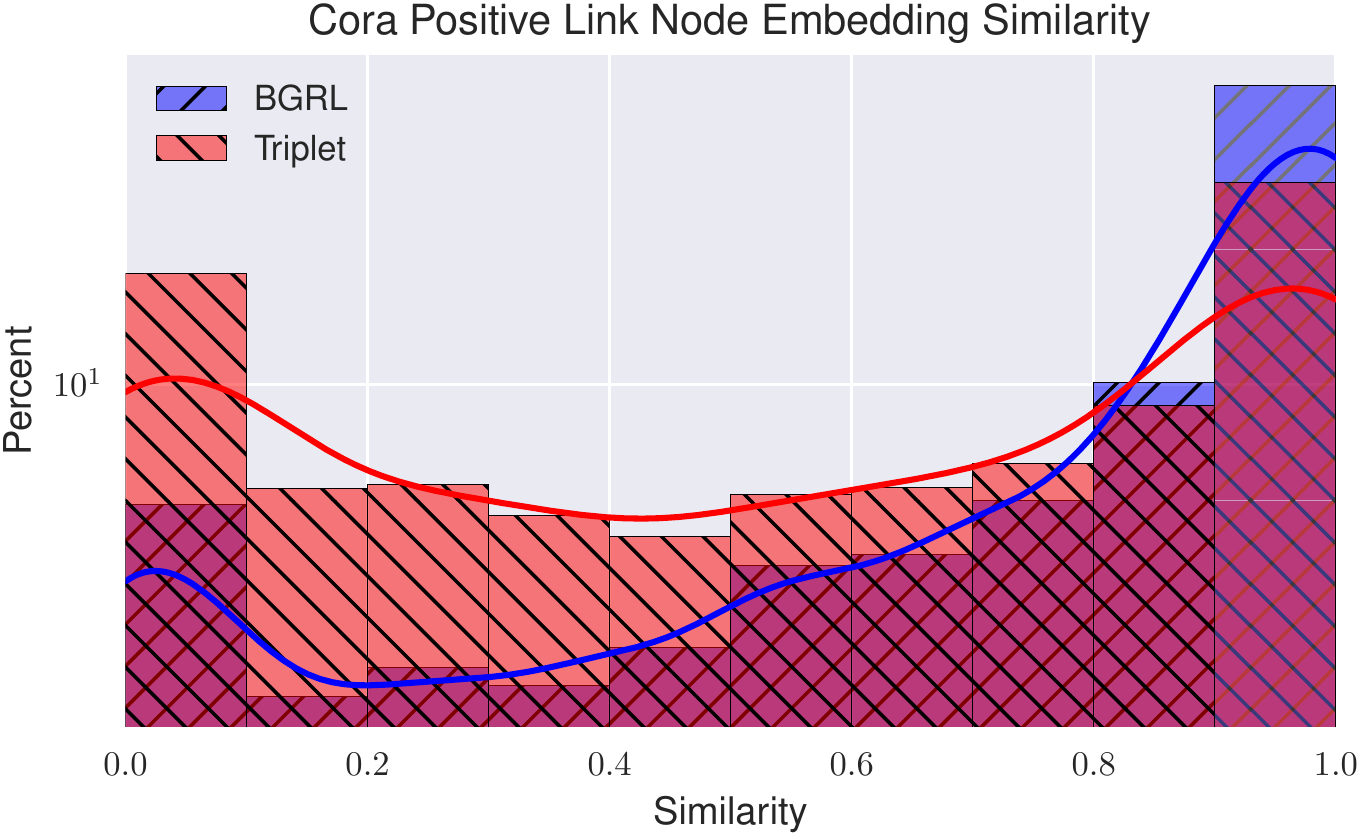}\label{fig:cora_ind_pos_dist}}
    \caption{\revAdd{These plots show similarities between node embeddings on \dset{Cora}. \textbf{Left:} distribution of similarity to \textit{non-neighbors} for \methodname{} and BGRL (closer to 0 is better). \textbf{Right:} distribution of similarity to \textit{neighbors} for \methodname{} and BGRL (closer to 1 is better). Note that the y-axis is on a logarithmic scale. \methodname{} clearly does a better job of ensuring that negative link representations are pushed far apart from those of positive links, but does not do as well at differentiating between positive links.}}
    \label{fig:cora_ind_dists}
\end{figure}

\end{document}